\pdfoutput=1
\documentclass[10pt,a4paper]{article}

\usepackage[margin=2.2cm]{geometry}
\usepackage[utf8]{inputenc}
\usepackage[T1]{fontenc}
\usepackage{lmodern}
\usepackage{amsmath,amssymb,amsfonts}
\usepackage{graphicx}
\usepackage{booktabs}
\usepackage{multirow}
\usepackage[table]{xcolor}
\definecolor{tablehl}{RGB}{238,246,252}
\usepackage{hyperref}
\usepackage[numbers,sort&compress]{natbib}
\usepackage{caption}
\usepackage{subcaption}
\usepackage{microtype}
\usepackage{setspace}

\definecolor{natureblue}{HTML}{1565C0}
\definecolor{natureteal}{HTML}{00897B}
\hypersetup{
    colorlinks=true,
    linkcolor=natureblue,
    citecolor=natureblue,
    urlcolor=natureteal
}

\newcounter{algocf}

\title{\Large\bfseries Laya-IsoToken: Day-One Clinical Codebook Switching via Closed-Form Operator Compilation}

\author{
    \normalsize Zeqiang Wang \\[0.45em]
    \small Department of Computer Science, University of Surrey, Guildford, UK \\
    \small \texttt{zeqiang.wang@surrey.ac.uk}
}

\date{}

\begin{document}

\maketitle

\begin{abstract}
Healthcare systems periodically revise diagnostic and procedural classifications (such as ICD-9 to ICD-10 and ICD-11), update hospital reimbursement groupers (MS-DRG), and deploy clinical coding across languages. Existing supervised coding networks bind clinical text encoders to fixed per-code classifier weights, forcing hospitals to collect and re-annotate tens of thousands of patient records before switching codebooks, whereas standalone large language models incur high latency and hallucinate invalid codes. Here we show that day-one clinical codebook switching becomes feasible without target-labeled clinical notes when fine-grained clinical evidence reading is decoupled from a strictly frozen multilingual concept space. Our framework, Laya-IsoToken, trains a local span reader once on source records with zero per-code classifier parameters, explicitly separating clinical text evidence from historical disease prevalence. Given only public descriptions of an unseen target codebook, closed-form matrix compilation aligns distributional shifts, compensates for paraphrased definitions, and deduplicates competing sibling codes in under three seconds, while a pre-deployment geometric residual routes unfamiliar concepts to offline language model verification. Across $32{,}804$ MIMIC-IV ICD-9 discharges, Laya-IsoToken matches published supervised baselines ($62.6\%$ Micro-F1) while reducing active calibration error by $32\%$. Switched with zero target training notes to $13{,}783$ unseen-patient ICD-10 discharges, MS-DRG reimbursement groups, and multilingual cohorts, it outperforms biomedical dense retrievers, semantic crosswalks ($+5.4$ F1 points on ICD-10, paired $d_z = 0.49$, $p < 0.001$), and official US government equivalence mappings ($+3.5$ points, $p < 0.001$) at $9.4\,\text{ms}$ per admission. Pairing this millisecond reader with an offline 31-billion-parameter verifier across $1{,}000$ full discharge summaries lifts zero-training ICD-10 coding to $40.6\%$ Micro-F1 ($57.0\%$ clinical category F1, $d_z = 0.66$, $p < 0.001$) while eliminating all $16.0\%$ invalid code hallucinations.
\end{abstract}

\section{Introduction}

Hospitals depend on standardized diagnostic, procedural, and reimbursement codebooks to fund inpatient care, adjust clinical risk, audit care quality, and track population disease burden~\cite{johnson2023mimic,johnson2023mimicnote,edin2023automated}. In clinical operations, these codebooks are never static: health ministries mandate nationwide migrations from ICD-9 to ICD-10 and ICD-11~\cite{boyd2013discriminatory,harrison2021icd11}, payers revise Diagnosis-Related Group (DRG) reimbursement rules annually, and regional institutions deploy coding across local languages and specialty catalogs. On day one of any such transition, a hospital possesses the new public codebook dictionary—code identifiers, official definitions, and hierarchy rules—but zero historical discharge summaries labeled under the new standard. Waiting months or years to accumulate and manually re-annotate tens of thousands of target admissions creates costly billing backlogs, breaks longitudinal cohort continuity, and places heavy pressure on scarce clinical coding staff.

Neither supervised medical coding networks nor standalone generative language models solve this day-one transition problem. Supervised label-attention architectures—including CAML~\cite{mullenbach2018explainable}, MultiResCNN~\cite{li2020multirescnn}, LAAT~\cite{vu2020label}, and PLM-ICD~\cite{huang2022plmicd,edin2023automated}—achieve fast, accurate coding within a single historical taxonomy, yet store medical knowledge inside millions of dedicated per-code classifier weights that become unusable whenever the catalog changes. Multi-version joint training~\cite{liu2026duallaat} improves rare-code recall across versions, but still requires labeled patient notes in every target schema prior to deployment. Human-curated government crosswalks such as the US Centers for Medicare and Medicaid Services General Equivalence Mappings (GEMs)~\cite{boyd2013discriminatory,boyd2015metrics} exist only for selected national transitions, suffer severe one-to-many branching across specialties, and cannot bridge reimbursement groupers or non-English catalogs. Conversely, generative large language models (LLMs) can read new code descriptions without retraining, yet standalone LLMs struggle with exact billing granularity and hallucinate invalid codes~\cite{soroush2024llm}, while multi-step agentic coders~\cite{edin2026symphony} require seconds to minutes per admission when screening tens of thousands of candidate codes.

A natural middle ground is to replace fixed per-code classifier rows with text embeddings of code descriptions~\cite{rios2018fewshot,song2021generalized,yuan2022synonyms}, yet conventional label-embedding coders still degrade sharply when switched to an unseen taxonomy. Two intuitive mechanisms explain this breakdown. First, allowing gradient descent to update the label encoder during source training bends the concept space around source-hospital phrasing and historical billing frequencies: in-domain accuracy rises slightly, but cross-version synonyms and cross-lingual equivalences drift apart, while post-hoc linear weight regression~\cite{heyden2026cast} shrinks decision boundaries toward the mean when projecting noisy per-code weights. Second, even when the concept space itself is preserved, switching to a new codebook alters the statistical environment of clinical matching in four predictable ways: non-English scripts or abbreviated billing titles shift the center and spread of code representations; reworded clinical definitions attenuate similarity relative to verbatim string matches; expanding or contracting catalog size shifts decision thresholds; and fine-grained sibling codes compete to claim the exact same phrase in a discharge summary.

\begin{figure*}[t]
    \centering
    \includegraphics[width=\textwidth]{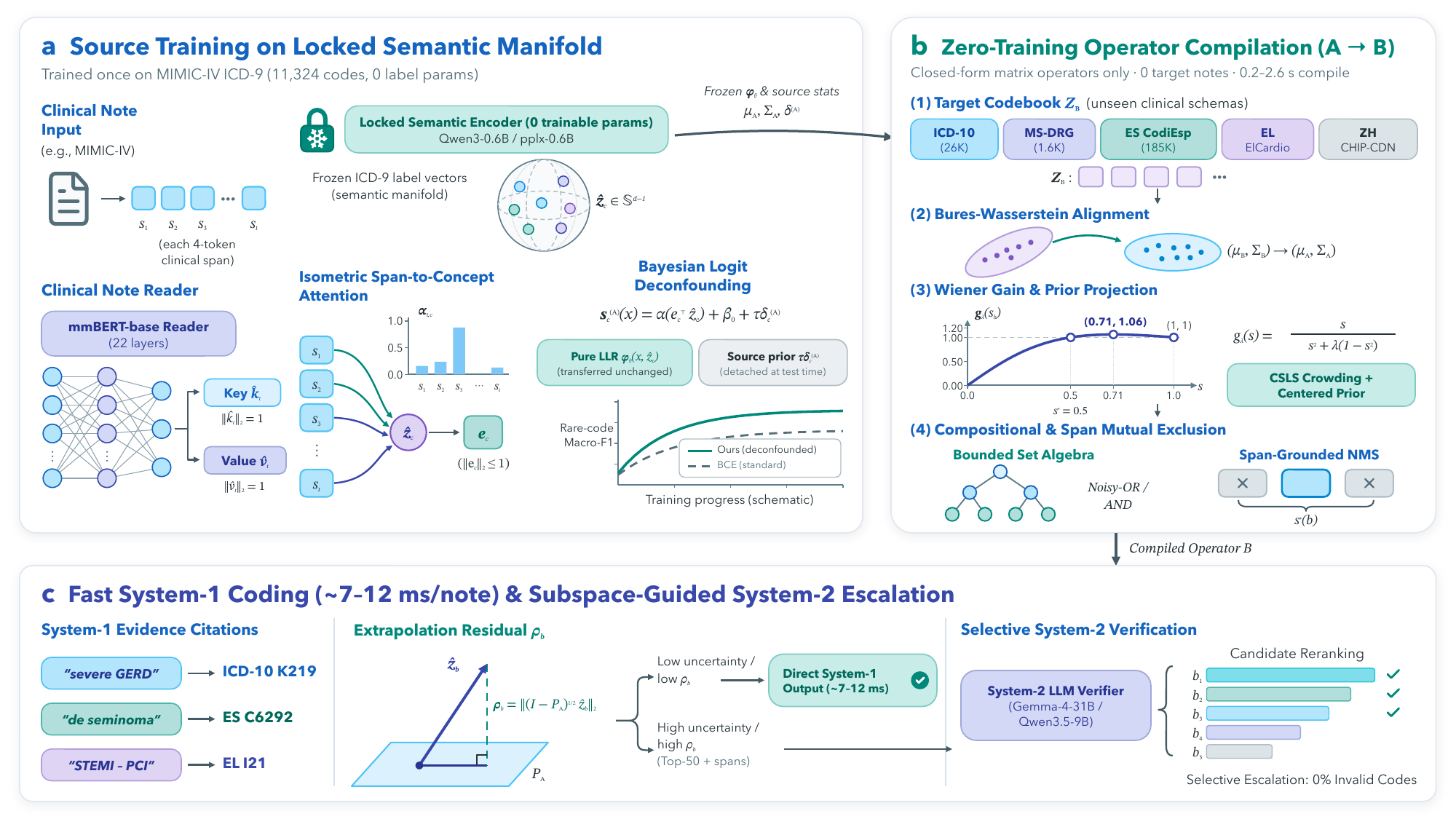}
    \caption{Overview of the Laya-IsoToken framework for day-one clinical codebook switching. \textbf{a}, One-time source training on MIMIC-IV ICD-9 pairs a clinical span reader (4-token local span pooling) against an offline locked codebook vectorizer (zero trainable label parameters), using Bayesian logit deconfounding ($\tau=0.5$) to separate clinical text evidence from historical disease prevalence. \textbf{b}, Offline codebook compilation transforms any unseen target catalog in $0.2$--$2.6\,\text{s}$ without target notes or gradient updates via Bures--Wasserstein distribution alignment, MMSE Wiener paraphrase gain $g_\lambda(s_b)$, whitened CSLS hubness centering, centered prior projection, bounded set algebra, and GPU-vectorized Span-Grounded Semantic Non-Maximum Suppression. \textbf{c}, Pre-deployment subspace extrapolation residual $\rho_b$ and causal 4-token citations drive selective escalation to offline System-2 LLM verifiers (\texttt{Gemma-4-31B-it} and \texttt{Qwen3.5-9B}).}
    \label{fig:framework}
\end{figure*}

To bridge this gap, we introduce Laya-IsoToken, a dual-process clinical coding framework that turns codebook switching from a months-long retraining project into a few-second offline compilation step (Fig.~\ref{fig:framework}). During one-time training on source hospital records, Laya-IsoToken pairs a fine-grained clinical span reader~\cite{marone2025mmbert} with a strictly frozen multilingual concept space~\cite{zhang2025qwen3embedding,perplexity2026pplxembed}—using zero trainable per-code parameters—and explicitly separates clinical text evidence from historical disease prevalence~\cite{menon2021longtail} so the reader learns portable clinical semantics rather than memorizing source billing rates. When switching to any unseen target codebook, Laya-IsoToken compiles the public code dictionary offline in 0.2--2.6 seconds without accessing target clinical notes. Five closed-form mathematical operations—orchestrated automatically from public codebook metadata—align cross-lingual and cross-format distributions~\cite{bhatia2019bures}, restore attenuated clinical paraphrases while keeping exact matches unchanged, suppress dense-catalog hubs~\cite{conneau2018word}, compose hierarchical procedure groups without neural hallucination, and enforce one-mention-one-code mutual exclusion directly on local text spans.

Freezing the concept space also provides pre-deployment safety awareness before a single target admission is coded. By measuring how far each unseen target code lies outside the source clinical subspace, Laya-IsoToken flags high-risk concepts in advance and drives a fast-and-slow clinical verification cascade~\cite{kahneman2011thinking}: the millisecond span reader proposes a catalog-valid shortlist anchored to verbatim clinical citations, while an offline open-weight LLM~\cite{qwen2026qwen35,google2026gemma4} audits ambiguous cases locally within hospital firewalls. Evaluated under TRIPOD+AI reporting standards~\cite{collins2024tripod+} across $32{,}804$ source MIMIC-IV ICD-9 discharges and five unseen target cohorts spanning ICD-10, MS-DRG reimbursement grouping, and Spanish, Greek, and Chinese terminologies ($115$ to $185{,}458$ codes), Laya-IsoToken matches supervised in-domain coding while establishing a fast, transparent, and open-source path for day-one clinical codebook portability.

\section{Results}
\label{sec:results}

To evaluate whether a single clinical reader can transition across real-world hospital coding regimes without retraining, we designed a six-cohort evaluation spanning three progressively harder operational axes (Supplementary Table~\ref{tab:cohorts}). First, we train the reader once on English MIMIC-IV ICD-9 discharge summaries and test whether removing all per-code classifier parameters sacrifices in-domain accuracy or probability calibration on $32{,}804$ held-out ICD-9 admissions ($15{,}425$ unseen patients). Second, freezing that source reader completely—with zero target clinical notes and zero target threshold tuning—we evaluate day-one switching across locked official test splits covering taxonomy expansion (MIMIC-IV ICD-10, $26{,}085$ diagnosis and procedure codes across $8{,}255$ source-disjoint patients), reimbursement task shift (single-label MIMIC-IV MS-DRG, $1{,}581$ groups across $7{,}340$ unseen patients), and cross-lingual coding (CLEF 2020 Spanish CodiEsp, $185{,}458$ codes plus $11{,}381$ compositional procedure groups), supplemented by two non-test boundary cohorts testing cross-script transfer (Greek-to-English ElCardioCC) and short-phrase normalization (Chinese CHIP-CDN). Third, we test whether the reader's pre-deployment geometric residual and millisecond span citations enable safe, accurate verification with offline open-weight LLMs across $1{,}000$ full-length ICD-10 discharge summaries.

\subsection{Eliminating per-code parameters preserves in-domain coding and improves calibration}
\label{sec:res_source}

\begin{figure*}[t]
    \centering
    \includegraphics[width=\textwidth]{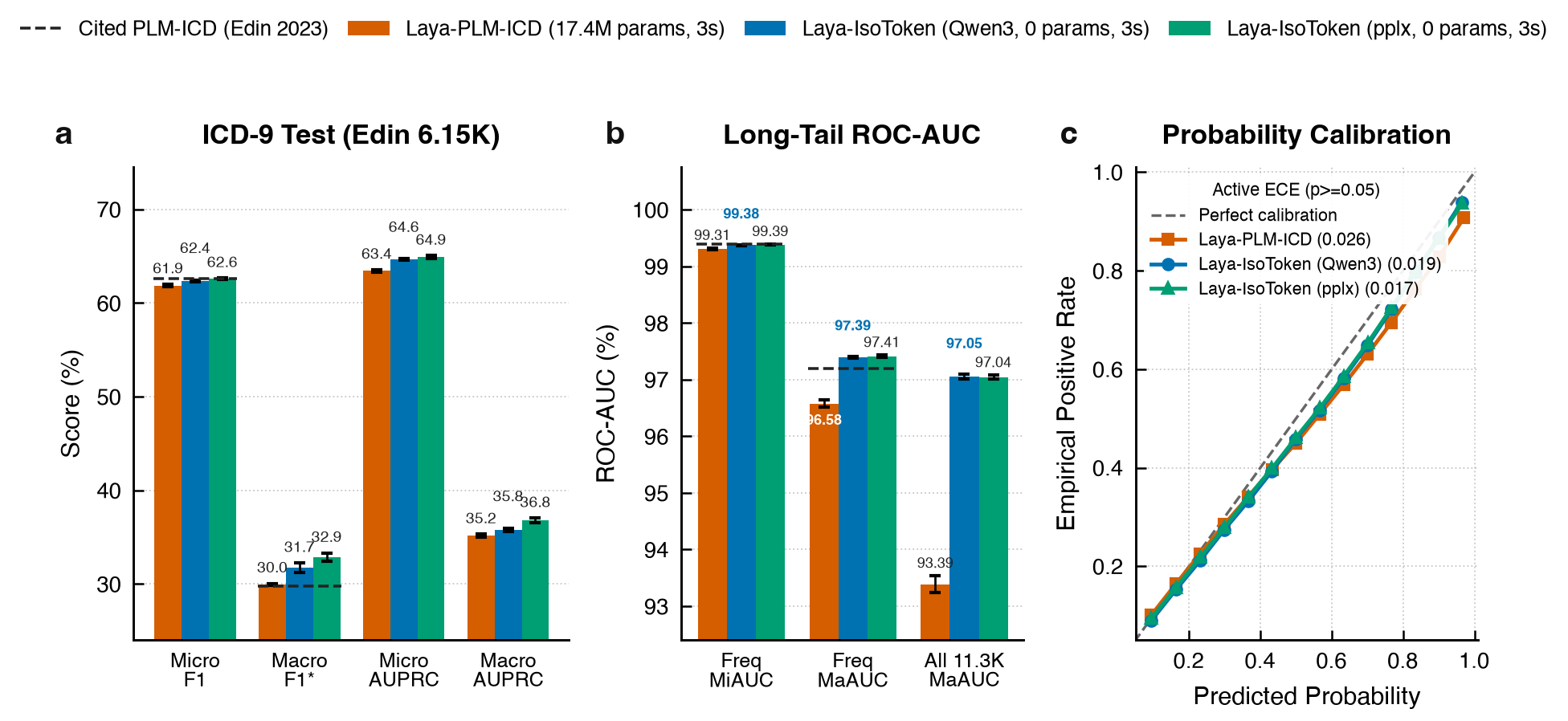}
    \caption{In-domain parity, long-tail discrimination, and probability calibration on the official locked MIMIC-IV ICD-9 test split ($n=32{,}804$ admissions, $15{,}425$ unseen patients, zero test-side threshold tuning). \textbf{a}, Exact Edin 6,150-code frequent-subset Micro~F1, deconfounded Macro~F1, Micro~AUPRC, and Macro~AUPRC comparing 3-seed 17.4M-parameter Laya-PLM-ICD against zero-label-parameter 3-seed Laya-IsoToken (\texttt{Qwen3} and \texttt{pplx} vectorizers); error bars denote $\pm 1$ SD and dashed lines mark published RoBERTa-PM PLM-ICD~\cite{edin2023automated}. \textbf{b}, Edin 6,150-code Micro and Macro ROC-AUC plus full 11,324-code Macro ROC-AUC showing a $+3.66$ point long-tail gain over Laya-PLM-ICD. \textbf{c}, 15-bin reliability diagram over active predictions ($p \ge 0.05$) showing reduced upper-tail overconfidence and $25\%$--$32\%$ lower Active ECE.}
    \label{fig:source_parity}
\end{figure*}

Removing all $17.4$ million trainable per-code weights entails no loss in single-taxonomy clinical coding; instead, zero-label-parameter Laya-IsoToken matches the strongest published supervised baseline and outperforms an architecture-matched per-code classifier across discrimination and calibration metrics on locked MIMIC-IV ICD-9 test admissions (Table~\ref{tab:source_test} and Fig.~\ref{fig:source_parity}). On Edin et al.'s exact $6{,}150$-code benchmark~\cite{edin2023automated}, three-seed Laya-IsoToken matches published RoBERTa-PM PLM-ICD in Micro~F1 ($62.60 \pm 0.06\%$ with the bidirectional \texttt{pplx} vectorizer; $62.36 \pm 0.10\%$ with causal \texttt{Qwen3}) while exceeding published Macro~F1 by $+3.1$ percentage points ($32.88 \pm 0.45\%$ versus $29.8\%$) at $7.05\,\text{ms}$ per admission on a single GPU. Against the controlled $17.4$M-parameter Laya-PLM-ICD baseline sharing the identical clinical span reader ($61.86 \pm 0.14\%$ Micro~F1), both frozen-codebook variants achieve consistent, statistically significant improvements ($+0.74\%$ and $+0.50\%$ Micro~F1, $p < 0.001$; paired $d_z = +0.08$ and $+0.04$, $55.2\%$ and $53.4\%$ patient win rates across $15{,}425$ patients; Supplementary Table~\ref{tab:effect_sizes}).

Decoupling clinical evidence from per-code weights yields its largest in-domain dividends in the long tail and in probability reliability (Fig.~\ref{fig:source_parity}b,c). Across the full $11{,}324$-code ICD-9 catalog—nearly half of which has fewer than ten training examples—sharing semantic geometry across code descriptions lifts Macro ROC-AUC by $+3.66$ percentage points over Laya-PLM-ICD ($97.05\%$ versus $93.39\%$). Simultaneously, whereas per-code classifiers exhibit pronounced upper-bin overconfidence on active predictions ($p \ge 0.05$), Laya-IsoToken reduces Active Expected Calibration Error~\cite{guo2017calibration} by $25.3\%$ (\texttt{Qwen3}) to $32.3\%$ (\texttt{pplx}), providing substantially better-calibrated probabilities before any cross-taxonomy transfer occurs.

\begin{table*}[t]
\centering
\caption{Official locked MIMIC-IV ICD-9 test benchmark ($n=32{,}804$ admissions, $15{,}425$ unseen patients, zero test-side threshold tuning). Published reference rows report supervised scores from Edin et al.~\cite{edin2023automated} on the identical split. Controlled models share the identical multilingual clinical span reader. For Laya-IsoToken models trained with Bayesian prior deconfounding ($\tau=0.5$), Micro~F1, ROC-AUC, AUPRC, and Active ECE ($p \ge 0.05$) evaluate the source-prior-calibrated score, whereas Macro~F1 reports the deconfounded pure evidence score. Boldface highlights the best zero-label-parameter model per column (excluding the 3.15M-parameter Unlocked control). All 3-seed Laya-IsoToken gains over matched 17.4M-parameter Laya-PLM-ICD are statistically significant under patient-clustered paired bootstrap ($p < 0.001$; paired $d_z = +0.08$ for \texttt{pplx} and $+0.04$ for \texttt{Qwen3}).}
\label{tab:source_test}
\scriptsize
\setlength{\tabcolsep}{2.8pt}
\resizebox{\textwidth}{!}{%
\begin{tabular}{@{}lrcccccccc@{}}
\toprule
Model & Label Params & \multicolumn{6}{c}{Exact Edin Frequent ($6{,}150$ codes)} & Active ECE & Latency \\
\cmidrule(lr){3-8} \cmidrule(lr){9-9} \cmidrule(lr){10-10}
 & & MiF1 $\uparrow$ & MaF1 $\uparrow$ & MiAUC $\uparrow$ & MaAUC $\uparrow$ & MiAUPRC $\uparrow$ & MaAUPRC $\uparrow$ & $p\ge0.05$ $\downarrow$ & ms/note $\downarrow$ \\
\midrule
\multicolumn{10}{@{}l}{\textit{Published supervised reference baselines on identical split (Edin et al.~\cite{edin2023automated})}} \\
CNN~\cite{edin2023automated} & Per-code & 52.4 & 12.6 & 98.1 & 89.4 & --- & --- & --- & --- \\
Bi-GRU~\cite{edin2023automated} & Per-code & 55.5 & 16.6 & 98.8 & 93.8 & --- & --- & --- & --- \\
CAML~\cite{mullenbach2018explainable} & Per-code & 58.6 & 19.3 & 98.8 & 90.7 & --- & --- & --- & --- \\
MultiResCNN~\cite{li2020multirescnn} & Per-code & 60.4 & 27.7 & 99.2 & 95.1 & --- & --- & --- & --- \\
LAAT~\cite{vu2020label} & Per-code & 61.7 & 26.4 & 99.3 & 96.0 & --- & --- & --- & --- \\
PLM-ICD (RoBERTa-PM)~\cite{huang2022plmicd} & Per-code & 62.6 & 29.8 & 99.4 & 97.2 & --- & --- & --- & --- \\
\midrule
\multicolumn{10}{@{}l}{\textit{Controlled clinical span reader (same 22-layer multilingual backbone, locked test)}} \\
Laya-PLM-ICD (3 seeds) & 17.4M & 61.86 $\pm$ 0.14 & 29.95 $\pm$ 0.11 & 99.31 $\pm$ 0.01 & 96.58 $\pm$ 0.07 & 63.44 $\pm$ 0.11 & 35.16 $\pm$ 0.15 & 0.0257 $\pm$ 0.0010 & 6.87 $\pm$ 0.01 \\
Laya-IsoToken-Unlocked (Qwen3, 3 seeds) & 3.15M & 62.69 $\pm$ 0.02 & 32.62 $\pm$ 0.22 & 99.45 $\pm$ 0.00 & 97.78 $\pm$ 0.04 & 65.19 $\pm$ 0.06 & 36.98 $\pm$ 0.22 & 0.0219 $\pm$ 0.0063 & 6.95 $\pm$ 0.09 \\
\rowcolor{tablehl} Laya-IsoToken (Qwen3 vectorizer, 3 seeds) & 0 & 62.36 $\pm$ 0.10 & 31.73 $\pm$ 0.51 & 99.38 $\pm$ 0.01 & 97.39 $\pm$ 0.02 & 64.65 $\pm$ 0.07 & 35.78 $\pm$ 0.18 & 0.0192 $\pm$ 0.0050 & 7.06 $\pm$ 0.02 \\
\rowcolor{tablehl} Laya-IsoToken (pplx vectorizer, 3 seeds) & 0 & \textbf{62.60 $\pm$ 0.06} & \textbf{32.88 $\pm$ 0.45} & \textbf{99.39 $\pm$ 0.01} & \textbf{97.41 $\pm$ 0.02} & \textbf{64.91 $\pm$ 0.15} & \textbf{36.80 $\pm$ 0.23} & \textbf{0.0174 $\pm$ 0.0037} & 7.05 $\pm$ 0.05 \\
\bottomrule
\end{tabular}%
}
\end{table*}

\subsection{Day-one codebook switching across taxonomies, reimbursement tasks, and languages}
\label{sec:res_transfer}

\begin{figure*}[t]
    \centering
    \includegraphics[width=\textwidth]{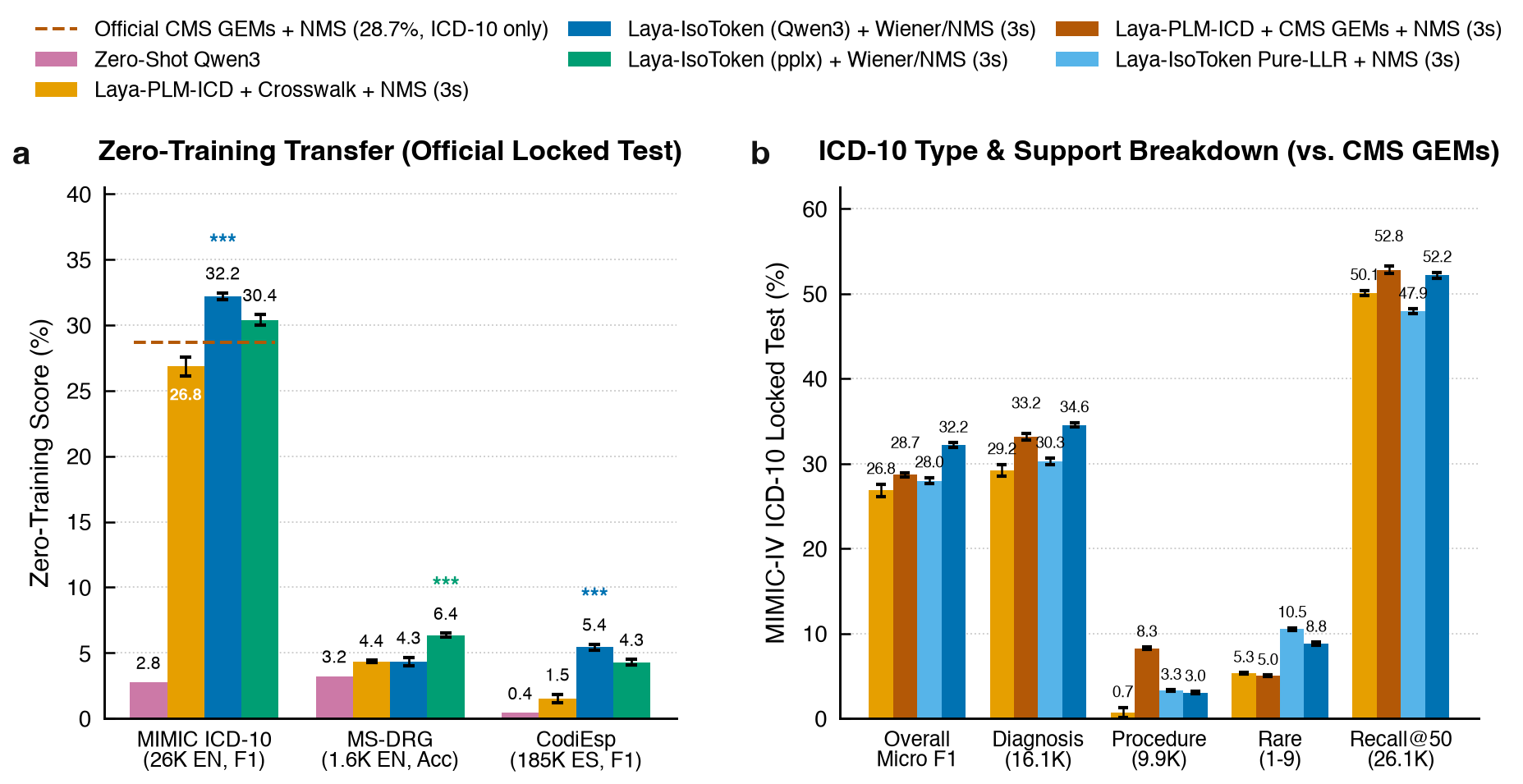}
    \caption{Zero-training clinical codebook transfer on official locked test cohorts (zero target training notes, zero target threshold tuning). \textbf{a}, Cross-version MIMIC-IV ICD-10 ($n=13{,}783$), cross-task MS-DRG ($n=11{,}869$), and cross-lingual Spanish CodiEsp ($n=250$) comparing Zero-Shot \texttt{Qwen3}, 3-seed Laya-PLM-ICD + Semantic Crosswalk + NMS, 3-seed Laya-IsoToken (\texttt{Qwen3} and \texttt{pplx} vectorizers), and official 2018 CMS GEMs + NMS (dashed reference, ICD-10 only); asterisks denote $p < 0.001$ under patient- or document-clustered paired bootstrap. \textbf{b}, Official locked MIMIC-IV ICD-10 breakdown comparing Crosswalk+NMS, official CMS GEMs+NMS, deconfounded Laya-IsoToken, and full compiled Laya-IsoToken across overall Micro~F1, diagnosis codes, PCS procedures, rare codes (support 1--9), and Recall@50.}
    \label{fig:transfer}
\end{figure*}

Compiled offline from public code descriptions alone under a single metadata-driven policy, the frozen source reader transfers directly to unseen clinical taxonomies in $0.2$--$2.6\,\text{s}$ and outperforms lexical, biomedical zero-shot bi-encoder, linear weight-transfer, semantic crosswalk, and human-curated government crosswalk baselines across target hospital workflows (Table~\ref{tab:transfer_test} and Fig.~\ref{fig:transfer}).

On locked patient-disjoint MIMIC-IV ICD-10 admissions ($n=13{,}783$, $26{,}085$ codes, compiled in $0.32\,\text{s}$ and scored in $9.41\,\text{ms}$/note), three-seed Laya-IsoToken reaches $32.20 \pm 0.26\%$ overall Micro~F1 and $50.82 \pm 0.37\%$ on 3-character clinical diagnosis categories. This represents a medium-to-large per-patient effect over the strongest trained baseline, Laya-PLM-ICD with Semantic Crosswalk and identical span deduplication ($+5.36$ F1 points, $p < 0.001$, paired $d_z = +0.49$, rank-biserial $r_{\text{rb}} = +0.45$, winning on $72.4\%$ of non-tied patients; $20/21$ clinical chapters in Supplementary Table~\ref{tab:chapters}), and a massive effect over biomedical zero-shot bi-encoders such as \texttt{MedCPT} and \texttt{BioLORD-2023-M} ($+29.95$ points over \texttt{BioLORD-2023-M}, $d_z = +1.80$, $98.0\%$ patient win rate; Supplementary Table~\ref{tab:effect_sizes}), confirming that global document-level pooling cannot resolve multi-diagnosis discharge narratives. Comparison against official human-curated 2018 US CMS General Equivalence Mappings (GEMs) exposes a complementary structural tradeoff: raw bidirectional GEMs collapse to $20.09\%$ Micro~F1 under one-to-many branching (up to $453$ target codes per ICD-9 code), and even after our span deduplication operator rescues GEMs to $28.67\%$, zero-crosswalk Laya-IsoToken wins overall by $+3.53$ points ($p < 0.001$). Notably, on rare ICD-10 codes (support 1--9), deconfounded Laya-IsoToken reaches $10.52 \pm 0.17\%$ Micro~F1 with zero target notes—exceeding published DualLAAT trained directly on $89{,}098$ ICD-10 notes ($8.4\%$) and nearly matching mixed ICD-9+10 supervision ($10.7\%$)~\cite{liu2026duallaat}. Conversely, human GEMs retain a clear advantage on synthetic 7-character ICD-10-PCS surgical codes ($8.25\%$ versus $3.05\%$); selectively applying public GEMs links only where branching fan-out is $\le 2$ raises zero-training procedure F1 $4.8\times$ to $14.66\%$ and overall ICD-10 Micro~F1 to $32.84 \pm 0.28\%$ ($p < 0.001$; Supplementary Table~\ref{tab:cal_hybrids} and Supplementary Fig.~\ref{fig:stage8_diag}b).

Switching from multi-label diagnosis coding to single-label hospital reimbursement on MIMIC-IV MS-DRG ($n=11{,}869$, $1{,}581$ uppercase labels, compiled in $0.21\,\text{s}$) yields $6.36 \pm 0.16\%$ exact Top-1 accuracy and $57.53\%$ Recall@50, outperforming Semantic Crosswalk ($+2.00$ points, $p < 0.001$, $r_{\text{rb}} = +0.25$, $62.4\%$ non-tie win rate) and lexical BM25 ($7.0\times$). Granularity auditing explains why exact text-only DRG accuracy appears modest: $91.4\%$ of admissions partition identical clinical conditions into structured billing severity tiers (with or without complication/comorbidity, CC/MCC) governed by secondary structured rules, causing exact accuracy to drop $2.6\times$ from the lowest to highest comorbidity quartile (Supplementary Table~\ref{tab:cal_hybrids}); collapsing CC/MCC suffixes into base clinical DRG families more than doubles three-seed Top-1 accuracy from $6.36 \pm 0.16\%$ to $14.73 \pm 0.25\%$ (versus $8.50 \pm 0.12\%$ for Semantic Crosswalk).

Cross-lingual evaluations delineate when neural concept compilation dominates versus when surface lexical matching suffices. On CLEF 2020 Spanish CodiEsp gold test cases ($185{,}458$ codes plus $11{,}381$ procedure groups, compiled in $2.58\,\text{s}$), Laya-IsoToken exceeds Semantic Crosswalk by $3.6\times$ ($5.44\%$ versus $1.51\%$, $p < 0.001$, paired $d_z = +0.54$, $r_{\text{rb}} = +0.73$, $86.2\%$ case win rate), and on cross-script Greek-to-English cardiology records (ElCardioCC), neural compilation retains $15.42\%$--$19.32\%$ Micro~F1 where lexical BM25 scores $0.00\%$. By contrast, when target inputs share verbatim native-language strings with catalog titles—particularly short diagnostic phrases in Chinese CHIP-CDN ($35.62\%$ BM25) or verbatim Spanish case phrases ($12.76\%$ BM25)—lexical matching outperforms English-trained span representations, and a zero-tuning union of BM25 with high-confidence neural anchors ($p \ge 0.50$) lifts CodiEsp to $12.94\%$ while doubling Top-50 recall to $24.8\%$ (Supplementary Table~\ref{tab:cal_hybrids}).

\begin{table*}[t]
\centering
\caption{Zero-training cross-taxonomy, cross-task, and cross-lingual transfer with zero target training notes and zero target threshold tuning. Trained transfer models share the identical source-trained clinical span reader; Qwen3 and pplx denote strictly offline codebook vectorizers compiled once per taxonomy ($0.2$--$2.6$\,s) under a single metadata-driven compilation policy. MIMIC-IV ICD-10 ($n=13{,}783$, $8{,}255$ unseen patients, $26{,}085$ codes), MIMIC-IV MS-DRG ($n=11{,}869$, $7{,}340$ unseen patients, $1{,}581$ labels), and Spanish CodiEsp ($n=250$, $185{,}458$ codes + $11{,}381$ groups) use locked official test splits; Greek ElCardioCC ($n=2{,}500$, $115$ labels) and Chinese CHIP-CDN ($n=2{,}000$, $37{,}645$ phrases) use locked non-test cohorts (official test labels withheld). Italic bottom rows provide published target-supervised reference points trained on labeled target-schema notes. Boldface highlights the best zero-training score per column (excluding the 3.15M-parameter Unlocked control). Values report mean $\pm$ SD across 3 seeds (\texttt{3s}).}
\label{tab:transfer_test}
\scriptsize
\setlength{\tabcolsep}{2.5pt}
\resizebox{\textwidth}{!}{%
\begin{tabular}{@{}lccccccccc@{}}
\toprule
Method (0 target notes unless noted) & \multicolumn{4}{c}{MIMIC ICD-10 Test (26K)} & \multicolumn{2}{c}{MS-DRG Test (1.6K)} & ES CodiEsp & EL Cardio & ZH CHIP \\
\cmidrule(lr){2-5} \cmidrule(lr){6-7} \cmidrule(lr){8-8} \cmidrule(lr){9-9} \cmidrule(lr){10-10}
 & MiF1 $\uparrow$ & Proc $\uparrow$ & Rare $\uparrow$ & R@50 $\uparrow$ & Top-1 $\uparrow$ & R@50 $\uparrow$ & Test MiF1 $\uparrow$ & MiF1 $\uparrow$ & MiF1 $\uparrow$ \\
\midrule
BM25 Lexical Baseline~\cite{robertson2009bm25} & 5.71 & 0.00 & 1.59 & 14.01 & 0.91 & 27.44 & \textbf{12.76} & 0.00 & \textbf{35.62} \\
ZeroShot MedCPT (biomedical bi-encoder)~\cite{jin2023medcpt} & 0.72 & 0.33 & 0.14 & 2.05 & 1.21 & 20.03 & 0.01 & 9.59 & 0.01 \\
ZeroShot BioLORD-2023-M (ontology bi-encoder)~\cite{remy2024biolord} & 2.25 & 1.08 & 1.12 & 6.76 & 3.44 & 48.98 & 0.64 & 12.02 & 0.96 \\
ZeroShot Qwen3-0.6B (bi-encoder)~\cite{zhang2025qwen3embedding} & 2.75 & 1.63 & 1.61 & 9.92 & 3.23 & 42.49 & 0.42 & 9.64 & 5.08 \\
Laya-PLM-ICD + CAST-Norm (3s)~\cite{heyden2026cast} & 4.41 $\pm$ 1.05 & 0.75 $\pm$ 0.05 & 1.14 $\pm$ 0.22 & 23.64 $\pm$ 2.02 & 1.95 $\pm$ 0.08 & 36.37 $\pm$ 0.94 & 0.36 $\pm$ 0.03 & 13.74 $\pm$ 1.19 & 0.24 $\pm$ 0.03 \\
Laya-PLM-ICD + CMS GEMs Forward (3s, ICD-10)~\cite{boyd2013discriminatory} & 25.69 $\pm$ 0.33 & 6.51 $\pm$ 0.13 & 4.87 $\pm$ 0.11 & \textbf{54.96 $\pm$ 0.37} & --- & --- & --- & --- & --- \\
Laya-PLM-ICD + CMS GEMs Bidir. (3s, ICD-10)~\cite{boyd2013discriminatory} & 20.09 $\pm$ 0.52 & 5.79 $\pm$ 0.29 & 4.56 $\pm$ 0.20 & 52.82 $\pm$ 0.42 & --- & --- & --- & --- & --- \\
Laya-PLM-ICD + CMS GEMs Bidir. + Span-NMS (3s) & 28.67 $\pm$ 0.25 & \textbf{8.25 $\pm$ 0.16} & 5.00 $\pm$ 0.10 & 52.82 $\pm$ 0.42 & --- & --- & --- & --- & --- \\
Laya-PLM-ICD + Semantic Crosswalk + NMS (3s) & 26.85 $\pm$ 0.73 & 0.72 $\pm$ 0.58 & 5.32 $\pm$ 0.08 & 50.09 $\pm$ 0.31 & 4.36 $\pm$ 0.08 & 50.02 $\pm$ 0.39 & 1.51 $\pm$ 0.33 & \textbf{20.00 $\pm$ 0.72} & 0.23 $\pm$ 0.09 \\
Laya-IsoToken-Unlocked (3.15M adapter, 3s) & 33.58 $\pm$ 0.42 & 3.96 $\pm$ 0.20 & 9.44 $\pm$ 0.21 & 53.73 $\pm$ 0.23 & 5.01 $\pm$ 0.44 & 55.65 $\pm$ 0.33 & 4.58 $\pm$ 0.22 & 14.45 $\pm$ 1.07 & 0.33 $\pm$ 0.01 \\
\midrule
Laya-IsoToken (Qwen3, deconfounded LLR + NMS, 3s) & 27.98 $\pm$ 0.37 & 3.26 $\pm$ 0.10 & \textbf{10.52 $\pm$ 0.17} & 47.92 $\pm$ 0.29 & --- & --- & --- & --- & --- \\
\rowcolor{tablehl} Laya-IsoToken (pplx vectorizer, 3s) & 30.39 $\pm$ 0.40 & 4.06 $\pm$ 0.49 & 9.33 $\pm$ 0.36 & 48.53 $\pm$ 0.48 & \textbf{6.36 $\pm$ 0.16} & \textbf{57.53 $\pm$ 1.70} & 4.29 $\pm$ 0.22 & 16.10 $\pm$ 3.44 & 0.90 $\pm$ 0.19 \\
\rowcolor{tablehl} Laya-IsoToken (Qwen3 vectorizer, full compiled, 3s) & \textbf{32.20 $\pm$ 0.26} & 3.05 $\pm$ 0.14 & 8.79 $\pm$ 0.20 & 52.16 $\pm$ 0.35 & 4.34 $\pm$ 0.32 & 52.93 $\pm$ 1.47 & 5.44 $\pm$ 0.23 & 15.42 $\pm$ 1.81 & 1.11 $\pm$ 0.40 \\
\midrule
\multicolumn{10}{@{}l}{\textit{Published target-supervised upper bounds (trained on labeled target-schema clinical notes; reference only)}} \\
\textit{Supervised PLM-ICD (89K ICD-10 notes, freq.)}~\cite{edin2023automated} & \textit{58.5} & --- & --- & --- & --- & --- & --- & --- & --- \\
\textit{Supervised DualLAAT (89K ICD-10 / +ICD-9, rare)}~\cite{liu2026duallaat} & --- & --- & \textit{8.4 / 10.7} & --- & --- & --- & --- & --- & --- \\
\textit{Supervised IAM System (750 CodiEsp notes, diag.)}~\cite{cossin2020iam} & --- & --- & --- & --- & --- & --- & \textit{68.7 (D)} & --- & --- \\
\bottomrule
\end{tabular}%
}
\end{table*}

\subsection{Why locked manifolds and closed-form operators govern portability}
\label{sec:res_ablation}

\begin{figure*}[t]
    \centering
    \includegraphics[width=\textwidth]{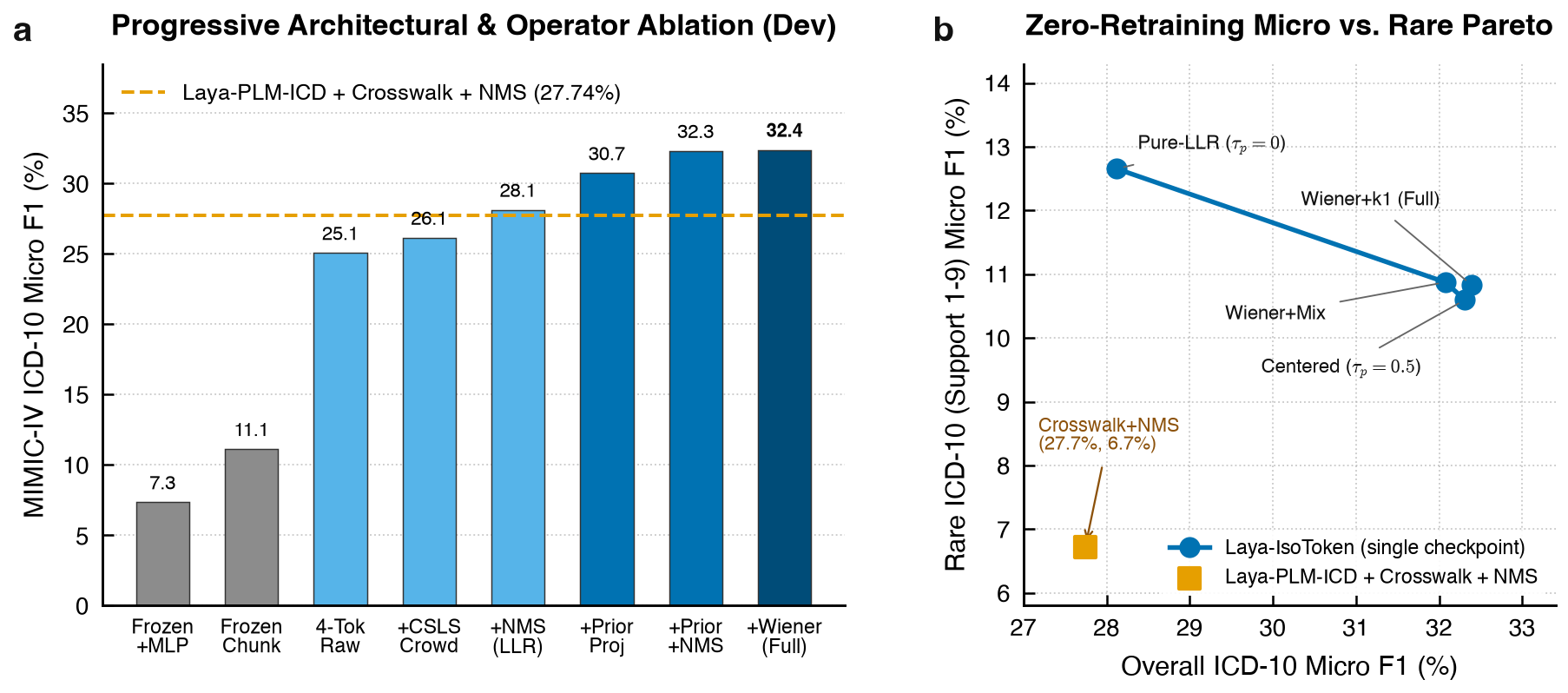}
    \caption{Component ablation and zero-retraining Pareto control on patient-disjoint MIMIC-IV ICD-10 dev ($n=9{,}077$, seed 42). \textbf{a}, Stepwise progression from frozen chunk-level baselines to 4-token span Laya-IsoToken and cumulative closed-form compilation operators applied to one frozen checkpoint. \textbf{b}, Frequent-versus-rare operating frontier traced by varying the single prior weight $\tau_{\text{proj}} \in \{0.0, 0.35, 0.50\}$ on one frozen checkpoint compared against Laya-PLM-ICD + Semantic Crosswalk + NMS.}
    \label{fig:ablation}
\end{figure*}

Controlled architectural and operator ablations isolate three causal mechanisms behind zero-training portability: locking the codebook manifold preserves cross-lingual geometry, scaling the offline codebook vectorizer improves target discrimination at zero runtime cost, and each closed-form operator corrects a distinct statistical distortion (Tables~\ref{tab:source_test}--\ref{tab:ablation}, Fig.~\ref{fig:ablation}, and Supplementary Table~\ref{tab:encoder_scaling}).

First, our three-seed unlocked-manifold control (\texttt{Laya-IsoToken-Unlocked}, inserting a $3.15$M-parameter zero-initialized residual adapter over \texttt{Qwen3} label vectors) exposes a fundamental specialization-versus-portability tradeoff. Allowing gradients to warp the label manifold adapts code vectors toward English MIMIC clinical phrasing—yielding modest gains on same-language English ICD-9 and ICD-10 ($+1.38$ F1 points on ICD-10, $33.58 \pm 0.42\%$ versus $32.20 \pm 0.26\%$)—but worsens source probability calibration ($0.0219$ versus $0.0192$ Active ECE) and systematically erodes non-English transfer across all three seeds, reducing Spanish CodiEsp by $15.8\%$ ($4.58 \pm 0.22\%$ versus $5.44 \pm 0.23\%$), Greek ElCardioCC ($14.45\%$ versus $15.42\%$), and cutting Chinese CHIP-CDN transfer by $70\%$ ($0.33 \pm 0.01\%$ versus $1.11 \pm 0.40\%$). Strictly locking the codebook at zero trainable label parameters preserves universal cross-lingual geometry and calibration, whereas upgrading the frozen offline vectorizer from \texttt{Qwen3-0.6B} ($1024$d) to \texttt{Qwen3-4B} ($2560$d; Supplementary Table~\ref{tab:encoder_scaling} and Supplementary Fig.~\ref{fig:encoder_scaling}) improves every locked test benchmark without enlarging the clinical note reader: ICD-10 rises to $33.01\%$ ($p < 0.001$), MS-DRG nearly doubles to $8.80\%$ exact Top-1 ($17.95\%$ base-family, paired $d_z = +0.15$, $r_{\text{rb}} = +0.44$, $72.1\%$ non-tie win rate), and Spanish CodiEsp climbs to $6.86\%$ ($d_z = +0.19$, $p = 0.002$).

Second, stepwise ablations on MIMIC-IV ICD-10 development notes ($n=9{,}077$; Table~\ref{tab:ablation} and Fig.~\ref{fig:ablation}a) separate representation resolution from post-training compilation. Replacing coarse 256-token chunk pooling ($11.12\%$) with 4-token local span attention over an external semantic vectorizer drives the primary jump to $30.31\%$, because distinct diagnoses within dense discharge summaries require sub-sentence spatial resolution. Applied to a single frozen prior-deconfounded checkpoint ($25.06\%$ raw cosine), the closed-form operators contribute additively with zero gradient updates: whitened CSLS hubness centering suppresses dense-catalog hubs ($+1.08$ points), span-grounded non-maximum suppression eliminates $42.5\%$ of same-mention sibling false positives ($+1.98$ points), centered Bayesian prior projection continuously traces the frequent-versus-rare Pareto curve between $12.65\%$ rare-code F1 and $32.39\%$ overall F1 without retraining (Fig.~\ref{fig:ablation}b), and MMSE Wiener spherical gain ($\lambda = 1/3$; Supplementary Fig.~\ref{fig:stage8_diag}a) selectively restores paraphrased diagnoses such as hypertension (\texttt{I10}) and reflux (\texttt{K219}) by $+1.9$ to $+7.8$ F1 points while leaving verbatim matches (\texttt{Z794}, \texttt{N179}) invariant within $0.1$ point (Fig.~\ref{fig:extrapolation}b).

\begin{table*}[t]
\centering
\caption{Stepwise architectural and zero-retraining operator ablation on MIMIC-IV ICD-10 patient-disjoint dev ($n=9{,}077$, $26{,}085$ codes, seed 42). All zero-retraining operator steps (bottom six rows) act on the exact same frozen Laya-IsoToken checkpoint without gradient updates. Pure likelihood-ratio mode with Span-NMS maximizes rare-code F1 ($12.65\%$), while composing centered prior projection, Span-NMS, and Wiener spherical gain reaches $32.39\%$ Micro~F1 and $10.83\%$ Rare~F1 simultaneously.}
\label{tab:ablation}
\scriptsize
\setlength{\tabcolsep}{3.8pt}
\resizebox{\textwidth}{!}{%
\begin{tabular}{@{}lcccccccc@{}}
\toprule
Configuration & Src ICD-9 & MiF1 $\uparrow$ & Prec $\uparrow$ & Rec $\uparrow$ & Diag $\uparrow$ & Proc $\uparrow$ & Rare $\uparrow$ & R@50 $\uparrow$ \\
\midrule
Frozen clinical reader + MLP label head (self-encoded labels) & 16.8 & 7.34 & 7.53 & 7.16 & 7.80 & 0.00 & 0.04 & --- \\
Frozen clinical reader + aligned cosine head (self-encoded labels) & 35.2 & 10.88 & 12.53 & 9.62 & 12.30 & 0.88 & 0.42 & --- \\
Frozen clinical reader + 256-token chunk isometric head & 34.4 & 11.12 & 15.31 & 8.74 & 12.15 & 0.26 & 0.63 & --- \\
Laya-IsoToken (12 layers unfrozen, $\tau=0$, Qwen3 vectorizer) & 60.9 & 30.01 & 32.79 & 27.67 & 32.27 & 3.54 & 7.66 & --- \\
Laya-IsoToken (22 layers unfrozen, $\tau=0$, Qwen3 vectorizer) & 62.2 & 30.31 & 35.54 & 26.43 & 32.70 & 2.94 & 8.47 & 51.64 \\
\midrule
Laya-IsoToken (22 layers, prior-deconfounded $\tau=0.5$, raw cosine) & 62.2 & 25.06 & 22.71 & 27.95 & 27.36 & 6.32 & 11.95 & --- \\
\quad + CSLS hubness centering ($\alpha_{\text{csls}}=0.5$) & --- & 26.14 & 25.23 & 27.12 & 28.36 & 5.81 & 12.35 & --- \\
\quad + Span-grounded Semantic NMS ($\cos_{\text{nms}}=0.80$) & --- & 28.12 & 33.53 & 24.22 & 30.53 & 3.18 & 12.65 & 48.49 \\
\quad + Centered Bayesian prior projection ($\tau_{\text{proj}}=0.5$) & --- & 30.73 & 31.32 & 30.16 & 33.02 & 4.45 & 10.23 & 52.58 \\
\quad + Centered prior projection + Span-NMS & --- & 32.31 & 39.29 & 27.44 & 34.85 & 2.30 & 10.60 & 52.58 \\
\rowcolor{tablehl} \quad + MMSE Wiener spherical gain ($\lambda=1/3$, full compiled) & --- & \textbf{32.39} & 37.01 & 28.79 & \textbf{34.90} & 2.85 & 10.83 & 52.47 \\
\bottomrule
\end{tabular}%
}
\end{table*}

\subsection{Pre-deployment extrapolation risk and causal span grounding}
\label{sec:res_extrapolation}

\begin{figure*}[t]
    \centering
    \includegraphics[width=\textwidth]{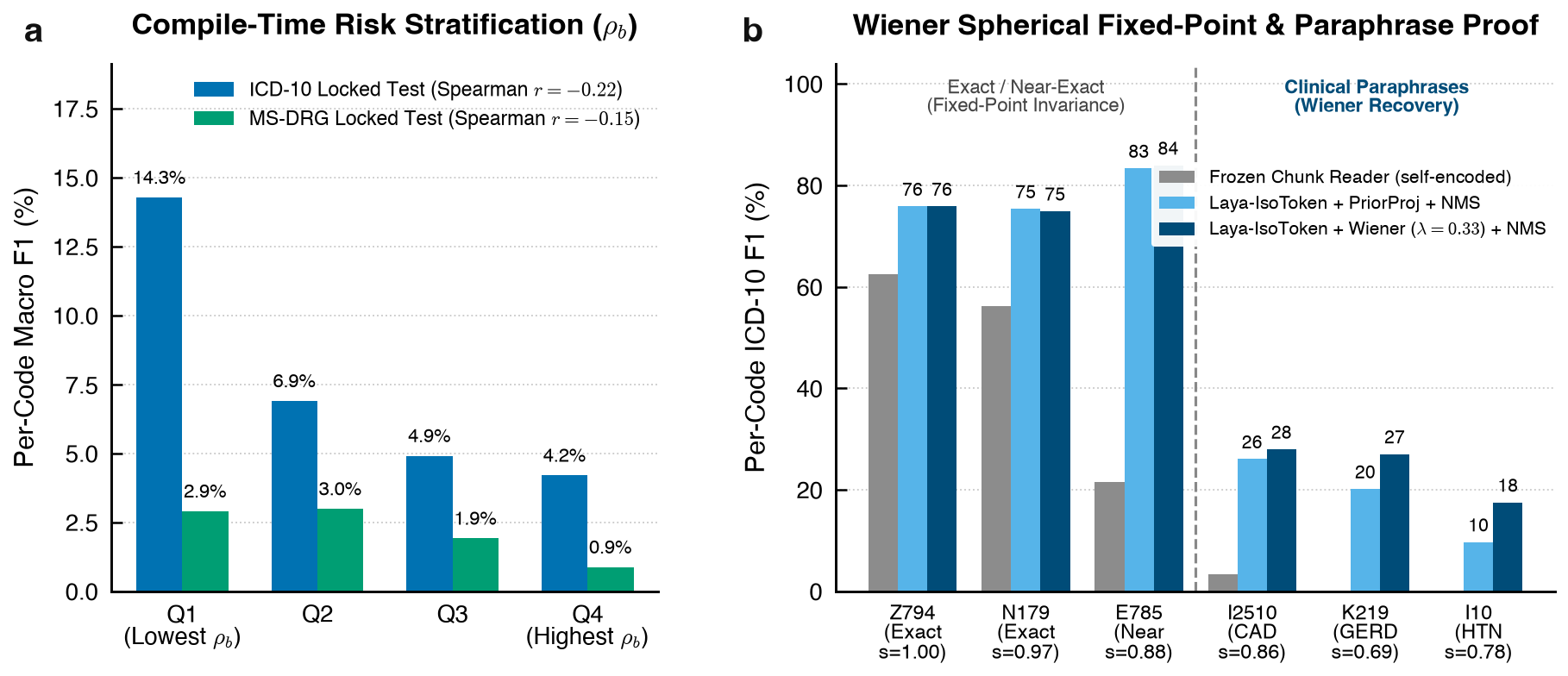}
    \caption{Compile-time subspace extrapolation diagnostics and Wiener paraphrase verification. \textbf{a}, Per-code Macro~F1 stratified by pre-deployment extrapolation residual quartiles $Q_1$--$Q_4$ on locked official MIMIC-IV ICD-10 test (Spearman $r=-0.22$, $14.3\% \to 4.2\%$) and locked MS-DRG test ($r=-0.15$). \textbf{b}, Per-code F1 verification of MMSE Wiener spherical gain $g_\lambda(s_b)$: exact lexical matches (\texttt{Z794}, \texttt{N179}, \texttt{E785}) remain invariant within $0.1$ F1 point, while clinically paraphrased codes (\texttt{I2510}, \texttt{K219}, \texttt{I10}) gain $+1.9$ to $+7.8$ F1 points without retraining.}
    \label{fig:extrapolation}
\end{figure*}

Before processing any target patient admission, the subspace extrapolation residual $\rho_b$ stratifies unseen target codes by out-of-manifold risk (Fig.~\ref{fig:extrapolation}a). Across locked MIMIC-IV ICD-10 test seeds, per-code Macro~F1 decays $3.4\times$ monotonically from the best-covered semantic quartile $Q_1$ ($14.28\%$) to the furthest quartile $Q_4$ ($4.21\%$, Spearman $r = -0.22$ to $-0.24$), with consistent monotonic decay on MS-DRG ($r = -0.15$ to $-0.17$). Ranking admissions by mean candidate extrapolation residual therefore provides a purely pre-deployment signal for routing unfamiliar cases to human or language-model review, outperforming margin-uncertainty hybrids without requiring labeled target calibration sets.

To test whether Laya-IsoToken's 4-token span citations reflect genuine causal evidence rather than post-hoc attention correlation, we performed a span-deletion and NegEx~\cite{chapman2001negex} clinical context audit across $6{,}248$ predicted codes in $500$ unseen-patient ICD-10 discharges (mean $742$ spans per note). Masking only the single highest-attended 4-token span ($0.13\%$ of note length) flips $33.69\%$ of predicted codes below threshold, whereas masking a random 4-token span flips $0.29\%$—establishing $117\times$ causal flip specificity across clinical concepts (\texttt{K219} $\to$ ``severe GERD'', MS-DRG \texttt{188} $\to$ ``with pleural effusions'', Spanish \texttt{C6292} $\to$ ``de seminoma cl\'{a}sico''). Contextual NegEx auditing further reveals that $10.25\%$ of remaining System-1 false positives sit inside local negation scopes (for example, ruled-out differential diagnoses), pinpointing sentence-level negation reasoning as the exact clinical capability where a slower language model verifier adds value.

\subsection{Fast-and-slow clinical verification cascade with open-weight LLMs}
\label{sec:res_cascade}

\begin{figure*}[t]
    \centering
    \includegraphics[width=\textwidth]{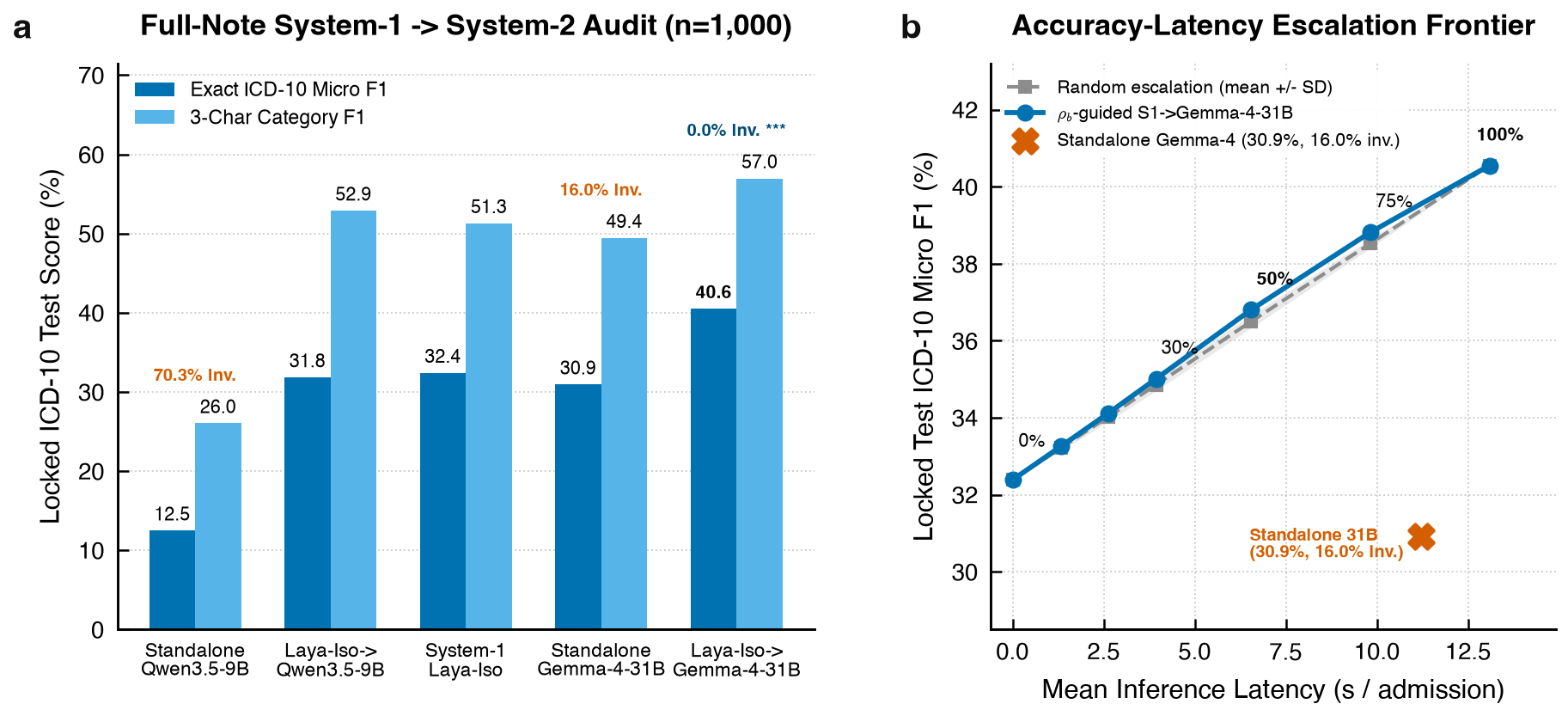}
    \caption{Full-note System-1 to System-2 clinical verification cascade on consecutive unseen-patient MIMIC-IV ICD-10 locked test admissions (\texttt{Gemma-4-31B-it}: $n=1{,}000$, $935$ unique patients; \texttt{Qwen3.5-9B}: $n=300$, $295$ unique patients; identical untruncated 4,000-word notes across every arm). \textbf{a}, Exact ICD-10 Micro~F1, 3-character diagnosis category F1, and invalid code hallucination rate across standalone LLMs, System-1 Laya-IsoToken alone, and System-1 $\to$ System-2 cascades; asterisks denote $p < 0.001$ ($d_z = 0.48$--$0.66$). \textbf{b}, Accuracy-latency Pareto frontier comparing pre-deployment $\rho_b$-guided selective escalation to \texttt{Gemma-4-31B-it} ($n=1{,}000$) against 50-replicate random escalation ($\pm 1$ SD band) and standalone \texttt{Gemma-4-31B-it}.}
    \label{fig:cascade}
\end{figure*}

Coupling millisecond System-1 span coding ($9.4\,\text{ms}$ GPU, $0.0\%$ invalid codes) with offline open-weight System-2 verifiers across untruncated 4,000-word MIMIC-IV ICD-10 discharge summaries delivers large clinical coding gains, eliminates invalid code hallucinations, and defines a clear model-capacity threshold for LLM adjudication (Table~\ref{tab:cascade} and Fig.~\ref{fig:cascade}).

Evaluated across $1{,}000$ consecutive unseen-patient admissions ($935$ unique patients), System-1 Laya-IsoToken alone ($32.40\%$ Micro~F1) already surpasses standalone zero-shot \texttt{Gemma-4-31B-it} ($30.93\%$ Micro~F1, $p = 0.007$) while running $1{,}190\times$ faster and avoiding the standalone LLM's $16.0\%$ hallucinated/non-existent code rate. When System-1 supplies its above-threshold proposals, remaining Top-50 shortlist candidates ($47.1\%$ shortlist recall), extrapolation residuals, and 4-token evidence spans to \texttt{Gemma-4-31B-it} for full structured audit, zero-training coding jumps to $40.55\%$ Micro~F1, $56.96\%$ 3-character clinical category F1, and $7.50\%$ PCS procedure F1 ($2.8\times$ System-1 alone) with $0.0\%$ invalid codes. Patient-clustered paired analyses confirm large standardized effect sizes over both System-1 alone ($+8.16$ points, $p < 0.001$, paired $d_z = +0.66$, $r_{\text{rb}} = +0.59$, winning on $79.4\%$ of patients) and standalone \texttt{Gemma-4-31B-it} ($+9.62$ points, $p < 0.001$, paired $d_z = +0.48$, $r_{\text{rb}} = +0.41$, $70.7\%$ patient win rate). Transition auditing (Supplementary Fig.~\ref{fig:stage8_diag}c) shows that the 31B verifier retains $89.4\%$ of System-1 true positives, rejects $55.2\%$ of false positives—including $58.2\%$ of negated-window errors—and rescues $1{,}732$ additional true positives from the candidate shortlist, tripling recall on paraphrased chronic conditions (\texttt{I10}, \texttt{I2510}, \texttt{K219}). Under compute budgets from $10\%$ to $75\%$, $\rho_b$-guided selective escalation consistently dominates random routing (Fig.~\ref{fig:cascade}b), capturing over half the full-audit gain while auditing only half of admissions ($36.81\%$ at $6.5\,\text{s}$ mean latency).

Repeating the identical full-note protocol with a compact 9-billion-parameter verifier (\texttt{Qwen3.5-9B}, $n=300$) establishes why System-1 grounding is indispensable and where verifier scale matters. Standalone \texttt{Qwen3.5-9B} collapses on the $26{,}085$-code catalog ($12.48\%$ Micro~F1, $70.3\%$ invalid codes); System-1 grounding eliminates all hallucinations and lifts performance $2.5\times$ to $31.82\%$ under confidence-anchored verification ($p \ge 0.50$). While wholesale delegation to a 9B model slightly trails System-1 alone ($30.73\%$ versus $31.98\%$), selectively escalating only the $20\%$ highest-$\rho_b$ admissions lifts even the 9B verifier above System-1 ($32.62\%$ at $1.79\,\text{s}$/note), demonstrating safe accuracy-latency control across hardware tiers.

\begin{table*}[t]
\centering
\caption{Full-note System-1 $\rightarrow$ System-2 clinical verification cascade on consecutive unseen-patient MIMIC-IV ICD-10 locked test admissions (Gemma-4-31B-it: $n=1,000$, $935$ unique patients; Qwen3.5-9B: $n=300$, $295$ unique patients; zero target training notes; identical untruncated 4,000-word discharge summaries across every arm). System-1 Laya-IsoToken ($9.4$\,ms GPU) constrains System-2 to valid catalog codes with grounded 4-token spans, eliminating invalid code hallucinations (16.0\%--70.3\% $\to 0.0\%$). With Gemma-4-31B-it, Full Audit achieves $40.55\%$ Micro~F1 ($56.96\%$ 3-char category F1, $7.50\%$ PCS procedure F1), outperforming both System-1 alone ($+8.16\%$, 95\% CI $[+7.57\%, +8.77\%]$, paired $d_z=+0.66$, $p<0.001$) and standalone Gemma-4-31B-it ($+9.62\%$, 95\% CI $[+8.69\%, +10.53\%]$, paired $d_z=+0.48$, $p<0.001$).}
\label{tab:cascade}
\scriptsize
\setlength{\tabcolsep}{3.4pt}
\resizebox{\textwidth}{!}{%
\begin{tabular}{@{}lcccccccc@{}}
\toprule
Routing Arm & MiF1 $\uparrow$ & 3-Char $\uparrow$ & Proc $\uparrow$ & Prec $\uparrow$ & Rec $\uparrow$ & Invalid $\downarrow$ & Latency (s) $\downarrow$ & LLM \\
\midrule
\multicolumn{9}{@{}l}{\textit{Compact open-weight System-2 verifier: Qwen3.5-9B ($n=300$, 295 unseen patients, untruncated notes)}} \\
Standalone Qwen3.5-9B (zero-shot) & 12.48 & 26.03 & 0.00 & 17.22 & 9.79 & 70.3\% & 18.73 & 100\% \\
Laya-IsoToken $\rightarrow$ Qwen3.5-9B Full Audit & 30.73 & 50.72 & 4.90 & 30.59 & 30.88 & 0.0\% & 8.83 & 100\% \\
Laya-IsoToken $\rightarrow$ Qwen3.5-9B Confidence-Anchored ($p\ge 0.50$) & 31.82 & 52.87 & 5.00 & 29.32 & 34.79 & 0.0\% & 8.83 & 100\% \\
Extrapolation-Guided $\rho_b$ $\rightarrow$ Qwen3.5-9B (20\% escalated) & 32.62 & 52.53 & 3.19 & 35.68 & 30.04 & 0.0\% & 1.79 & 20\% \\
\midrule
\multicolumn{9}{@{}l}{\textit{Frontier open-weight System-2 verifier: Gemma-4-31B-it ($n=1,000$, 935 unseen patients, untruncated notes)}} \\
System-1 Laya-IsoToken Alone (0 LLM calls) & 32.40 & 51.28 & 2.72 & 36.46 & 29.15 & 0.0\% & 0.01 & 0\% \\
Standalone Gemma-4-31B-it (zero-shot) & 30.93 & 49.42 & 2.94 & 42.12 & 24.43 & 16.0\% & 11.20 & 100\% \\
Extrapolation-Guided $\rho_b$ $\rightarrow$ Gemma-4-31B (30\% escalated) & 35.01 & 53.13 & 3.65 & 39.45 & 31.47 & 0.0\% & 3.93 & 30\% \\
Extrapolation-Guided $\rho_b$ $\rightarrow$ Gemma-4-31B (50\% escalated) & 36.81 & 54.36 & 4.32 & 41.28 & 33.21 & 0.0\% & 6.54 & 50\% \\
Extrapolation-Guided $\rho_b$ $\rightarrow$ Gemma-4-31B (75\% escalated) & 38.82 & 55.69 & 6.01 & 43.59 & 34.98 & 0.0\% & 9.81 & 75\% \\
Laya-IsoToken $\rightarrow$ Gemma-4-31B Confidence-Anchored ($p\ge 0.50$) & 38.79 & 56.57 & 7.45 & 38.65 & 38.93 & 0.0\% & 13.08 & 100\% \\
\rowcolor{tablehl} Laya-IsoToken $\rightarrow$ Gemma-4-31B Full Audit & \textbf{40.55} & \textbf{56.96} & \textbf{7.50} & 44.71 & 37.10 & 0.0\% & 13.08 & 100\% \\
\bottomrule
\end{tabular}%
}
\end{table*}

\section{Discussion}
\label{sec:discussion}

Hospital coding standards evolve faster than supervised neural classifiers can be relabeled, retrained, and revalidated across health systems. Nationwide migrations from ICD-9 to ICD-10 and ICD-11, annual updates to MS-DRG reimbursement tables, and multilingual hospital deployments routinely invalidate closed-set heads wired to historical label indices. Our findings show that decoupling fine-grained clinical span reading from mutable per-code parameters—and compiling new public codebooks via closed-form matrix operations on a frozen multilingual concept manifold—resolves this deployment bottleneck. Trained once on source records, Laya-IsoToken matches published supervised coding on MIMIC-IV ICD-9 while cutting active calibration error by up to $32\%$, switches to unseen $1.6$K--$185$K codebooks in under three seconds without target clinical notes, and acts as a millisecond citation-generating front-end that boosts an offline 31B verifier by $+9.62$ F1 points ($d_z = +0.48$) while eliminating invalid code hallucinations.

Four mechanistic insights explain why representation-level compilation outperforms both post-hoc weight projection and static crosswalks. First, strictly locking the codebook vectorizer is what preserves cross-lingual portability and probability calibration: our three-seed unlocked adapter control confirms that allowing gradient updates on label embeddings buys a small same-language English gain ($+1.38$ points on ICD-10) by fitting source hospital phrasing, at the cost of worsening calibration and eroding Spanish, Greek, and Chinese transfer ($-15.8\%$ to $-70.3\%$). Conversely, upgrading only the frozen offline vectorizer to 4B parameters improves ICD-10, doubles MS-DRG accuracy, and boosts Spanish coding at zero inference overhead. Second, Bayesian prior deconfounding separates clinical text evidence from historical disease prevalence during training, enabling a single frozen checkpoint to match supervised rare-code ICD-10 DualLAAT ($10.52\%$ versus $8.4\%$) with zero target notes. Third, closed-form compilation fixes distinct, measurable geometric shifts—optimal transport aligns non-Latin scripts and uppercase billing catalogs, whitened hubness scaling suppresses dense-catalog hubs, GPU span deduplication enforces one-mention-one-code mutual exclusion in $0.34\,\text{ms}$, and Wiener spherical gain selectively recovers paraphrased clinical definitions. Fourth, pre-deployment subspace residuals ($\rho_b$) identify out-of-manifold risk before any target patient note is seen, turning System-1 and System-2 coding into a controllable, privacy-preserving hospital workflow.

Comparisons with official US CMS General Equivalence Mappings and lexical BM25 clarify how Laya-IsoToken integrates with existing health-informatics tools. Human-curated crosswalks remain valuable for synthetic 7-character ICD-10-PCS surgical syntax where multi-axial procedure grammar diverges from narrative prose, yet unconstrained crosswalks suffer severe branching precision loss and do not exist for reimbursement groupers, non-English catalogs, or future ICD-11 transitions. Restricting public GEMs strictly to low-branching surgical links ($\text{fan-out} \le 2$) combines expert procedural rules with neural diagnosis coding ($32.84\%$ overall, $14.66\%$ PCS). Similarly, while semantic span compilation outperforms BM25 by $5.6\times$--$7.0\times$ on narrative English discharge summaries and cross-script Greek records, exact lexical matching excels when target inputs mirror catalog strings verbatim (Chinese short phrases and verbatim Spanish case reports), where combining BM25 with high-confidence Bayesian neural anchors ($p \ge 0.50$) captures both lexical precision and semantic recall.

\paragraph{Limitations.}
Four boundaries define concrete priorities for clinical translation. First, although dual-process verification reaches $40.55\%$ exact Micro~F1 and $56.96\%$ 3-character clinical category F1 with zero target training notes, day-one zero-training coding remains below fully supervised models trained on tens of thousands of target-labeled records ($58.5\%$ frequent-code F1~\cite{edin2023automated}), particularly in structural chapters such as Injury/Trauma ($7.8\%$), Obstetrics ($8.9\%$), and 7-character PCS procedures, motivating rapid few-shot adaptation once initial post-transition records accumulate. Second, $10.3\%$ of System-1 false positives arise inside local clinical negation windows; although the 31B verifier rejects $58.2\%$ of these errors, lightweight negation-aware span gating could suppress ruled-out diagnoses directly at millisecond speed. Third, fine-tuning transformer layers solely on English discharge summaries introduces mild cross-lingual drift on non-English notes, suggesting multilingual self-distillation during source training. Finally, real-world MS-DRG grouping combines clinical narratives with structured age, sex, discharge disposition, and secondary CC/MCC complication tables; fusing structured electronic health record variables with Laya-IsoToken and validating prospectively on multi-center ICD-11 transitions represent essential next steps.

\section{Methods}

\subsection{Problem Formulation, Data Isolation, and TRIPOD+AI Reporting}

Let $\mathcal{D}_A = \{(x_i, Y_i^{(A)})\}_{i=1}^{N_A}$ denote a labeled source clinical corpus annotated under source codebook $A = \{a_1, \dots, a_{|A|}\}$ with official textual descriptions $\{t_a\}_{a \in A}$, and let $B = \{b_1, \dots, b_{|B|}\}$ denote an unseen target classification schema defined solely by public code descriptions $\{t_b\}_{b \in B}$ and optional public compositional group relations (such as hierarchical procedure prefixes). Our objective is to train a clinical evidence reader $\phi_\theta$ once on $\mathcal{D}_A$ and, given only the public specification of $B$, compile a target scoring operator $\mathcal{C}(\phi_\theta, \mathcal{S}_A, B)$ using closed-form matrix operations with zero target clinical documents, zero target annotations, and zero gradient updates, where $\mathcal{S}_A$ stores only aggregate source codebook statistics ($\mu_A, \Sigma_A, Z_A, \delta^{(A)}$). Reporting follows the TRIPOD+AI statement~\cite{collins2024tripod+}, including explicit cohort reconciliation, probability calibration, subgroup fairness analysis, and open-source code availability.

We enforce a strict two-tier data isolation protocol throughout the study. First, all architectural choices, loss formulations, and compilation hyperparameters were fixed exclusively on source MIMIC-IV ICD-9 validation data and non-test target development splits; all official test splits (MIMIC-IV ICD-9 test, MIMIC-IV ICD-10 test, MIMIC-IV MS-DRG test, and CLEF 2020 Spanish CodiEsp gold test) remained sealed until checkpoints and compilation constants were locked, after which a single zero-tuning evaluation pass was executed using frozen source-validation probability thresholds. Second, for every cross-schema evaluation within MIMIC-IV~\cite{johnson2023mimic,johnson2023mimicnote}, we enforce strict patient-level disjointness by excluding every admission whose \texttt{subject\_id} appears in the MIMIC-IV ICD-9 source training or validation partitions (removing $6{,}019$ overlapping admissions from Edin et al.'s $19{,}802$ ICD-10 test admissions to yield $13{,}783$ unseen-patient admissions across $8{,}255$ unique patients). Two additional cross-lingual benchmarks whose official test labels are withheld by challenge organizers (Greek ElCardioCC, $N = 2{,}500$ train cohort; Chinese CHIP-CDN, $N = 2{,}000$ dev cohort) are evaluated strictly as frozen non-test boundary cohorts. All clinical notes follow standardized preprocessing capped at $4{,}000$ whitespace words, segmented into up to $32$ chunks of $256$ tokens using the pinned multilingual \texttt{mmBERT-base} tokenizer (\texttt{convaiinnovations/laya-multilingual})~\cite{marone2025mmbert}.

\subsection{Laya-IsoToken Architecture}

Conventional multi-label medical coders~\cite{mullenbach2018explainable,vu2020label,huang2022plmicd} learn a $d \times |A|$ classification matrix whose rows cannot transfer to new codes without retraining or lossy linear regression, while earlier zero-shot coding architectures~\cite{rios2018fewshot,song2021generalized,yuan2022synonyms} learn trainable label encoders that overfit source-codebook lexical phrasing. Laya-IsoToken pairs a multilingual clinical span reader with zero trainable per-code parameters by grounding span-level evidence reading on an offline locked semantic manifold (Fig.~\ref{fig:framework}a).

\paragraph{Offline Locked Codebook Vectorizer.}
Given a frozen pretrained text vectorizer $f_{\text{sem}}$ with output dimension $d_z$ (causal last-token-pooled \texttt{Qwen3-Embedding-0.6B}~\cite{zhang2025qwen3embedding} at $d_z = 1024$ as primary vectorizer, diffusion-pretrained bidirectional mean-pooled \texttt{pplx-embed-v1-0.6B}~\cite{perplexity2026pplxembed} at $d_z = 1024$ as a three-seed independent architectural control, plus $22$M English-only \texttt{all-MiniLM-L6-v2}~\cite{wang2020minilm} at $d_z = 384$ and \texttt{Qwen3-Embedding-4B}~\cite{zhang2025qwen3embedding} at $d_z = 2560$ for dimension and capacity scaling), every code $c$ in any codebook is encoded once offline from its official public description into a raw vector $z_c \in \mathbb{R}^{d_z}$; $f_{\text{sem}}$ never runs at patient inference time. Source codebook vectors are centered by the source mean $\mu_A = \frac{1}{|A|}\sum_{a \in A} z_a$ and normalized onto the unit hypersphere $\mathbb{S}^{d_z - 1}$:
\begin{equation}
\hat{z}_a = \frac{z_a - \mu_A}{\max(\|z_a - \mu_A\|_2, \epsilon)}, \qquad a \in A,
\label{eq:unit_label}
\end{equation}
with $\epsilon = 10^{-6}$. To isolate the causal effect of locking the manifold, we also train a three-seed unlocked-label control (\texttt{Laya-IsoToken-Unlocked}) that passes normalized label prototypes through a shared $3.15$M-parameter residual MLP initialized at the identity map.

\paragraph{Fine-Grained 4-Token Clinical Span Reader.}
Each clinical note $x$ is encoded chunk-by-chunk by the 22-layer multilingual encoder \texttt{mmBERT-base} (\texttt{laya-multilingual}~\cite{marone2025mmbert}, $d_h = 768$), whose token embedding table remains frozen to preserve multilingual lexical alignment while its top transformer layers ($22$ layers in primary models; $12$ layers in ablation) and final normalization are fine-tuned. Valid non-special token states (masking \texttt{[CLS]}, \texttt{[SEP]}, and \texttt{[PAD]}) are mean-pooled into contiguous non-overlapping 4-token clinical spans, yielding $S \le 2048$ span states $h_t \in \mathbb{R}^{d_h}$ per admission ($64$-fold finer spatial resolution than 256-token chunk pooling). Two residual feedforward projectors map each span state into unit-normalized attention keys and semantic concept values on $\mathbb{S}^{d_z - 1}$:
\begin{equation}
\hat{k}_t = \frac{\psi_K(h_t)}{\max(\|\psi_K(h_t)\|_2, \epsilon)}, \qquad \hat{v}_t = \frac{\psi_V(h_t)}{\max(\|\psi_V(h_t)\|_2, \epsilon)}, \qquad t = 1, \dots, S,
\end{equation}
where each $\psi(\cdot)$ applies layer normalization, a linear skip map from $d_h$ to $d_z$, and a two-layer residual GELU feedforward block with dropout $0.1$.

\paragraph{Isometric Span-to-Concept Attention.}
For any candidate concept vector $\hat{z}_c \in \mathbb{S}^{d_z - 1}$, attention weights over valid spans are computed via scaled cosine similarity $\alpha_{t,c} = \operatorname{softmax}_t(\gamma_{\text{attn}} \hat{k}_t^\top \hat{z}_c)$ with learnable temperature $\gamma_{\text{attn}} \in [1, 100]$. The pooled clinical evidence vector is $e_c = \sum_{t=1}^{S} \alpha_{t,c} \hat{v}_t$, and its raw semantic evidence logit is:
\begin{equation}
\phi_\theta(x, \hat{z}_c) = \alpha_{\text{sc}} \, (e_c^\top \hat{z}_c) + \beta_0,
\label{eq:pure_llr}
\end{equation}
where $\alpha_{\text{sc}} \in [1, 100]$ and $\beta_0$ are global scalars shared across all codes. Leaving $e_c$ unnormalized preserves absent-concept contraction ($\|e_c\|_2 \le 1$): when no clinical span supports concept $c$, diverse span vectors cancel inside $e_c$ rather than amplifying noise onto the unit sphere. Because $e_c^\top \hat{z}_c = \sum_{t=1}^{S} \alpha_{t,c} (\hat{v}_t^\top \hat{z}_c)$, exact scoring factorizes into chunked matrix multiplications of complexity $O(B S C)$ rather than $O(B S C d_z)$, yielding single-digit millisecond latency and an exact argmax 4-token evidence span $s^*(c) = \arg\max_t \alpha_{t,c}$ for every code.

\paragraph{Bayesian Prior-Deconfounded Source Training.}
To prevent attention weights from memorizing source billing frequencies, we isolate empirical source prevalence during training via a frozen Jeffreys-smoothed empirical log-odds offset~\cite{menon2021longtail} $\delta_a^{(A)} = \operatorname{logit}(\pi_a^{(A)}) - \bar{\ell}_A$, where $\pi_a^{(A)} = \frac{n_a + 0.5}{N_A + 1}$ and $\bar{\ell}_A = \operatorname{logit}\!\big(\frac{1}{|A|}\sum_{a' \in A} \pi_{a'}^{(A)}\big)$:
\begin{equation}
s_a^{(A)}(x) = \phi_\theta(x, \hat{z}_a) + \tau \, \delta_a^{(A)},
\label{eq:bayes_deconfound}
\end{equation}
trained with full-codebook binary cross-entropy over all $|A| = 11{,}324$ source codes at $\tau = 0.5$. On in-domain codebook $A$, $s_a^{(A)}(x)$ ($\tau_{\text{eval}} = 0.5$) provides the prior-calibrated operating point and $\phi_\theta(x, \hat{z}_a)$ ($\tau_{\text{eval}} = 0$) provides the deconfounded Macro-F1 operating point; on migration to any unseen codebook $B$, $\tau \delta^{(A)}$ is discarded cleanly.

\subsection{Closed-Form Offline Codebook Compilation ($A \to B$)}

Given an unseen target codebook $B$ with raw description embeddings $Z_B \in \mathbb{R}^{|B| \times d_z}$, Laya-IsoToken compiles target scoring parameters offline in $0.2$--$2.6\,\text{s}$ without target clinical notes (Fig.~\ref{fig:framework}b and Algorithm~\ref{alg:isotoken}).

\paragraph{1. Moment and Bures--Wasserstein Optimal Transport Alignment.}
Non-Latin scripts and uppercase reimbursement catalogs exhibit mean shifts ($\mu_B \neq \mu_A$) and anisotropic covariance drift relative to $A$. Using regularized sample covariances $\Sigma_A^{\text{reg}} = (1 - \eta)\Sigma_A + \eta \frac{\operatorname{tr}(\Sigma_A)}{d_z} I$ and $\Sigma_B^{\text{reg}} = (1 - \eta)\Sigma_B + \eta \frac{\operatorname{tr}(\Sigma_B)}{d_z} I$ ($\eta = 0.05$), the symmetric positive-definite Monge optimal transport map minimizing $\mathcal{W}_2^2(\mathcal{N}(\mu_B, \Sigma_B^{\text{reg}}), \mathcal{N}(\mu_A, \Sigma_A^{\text{reg}}))$~\cite{bhatia2019bures} is:
\begin{equation}
T_{B \to A} = (\Sigma_B^{\text{reg}})^{-1/2} \Big( (\Sigma_B^{\text{reg}})^{1/2} \, \Sigma_A^{\text{reg}} \, (\Sigma_B^{\text{reg}})^{1/2} \Big)^{1/2} (\Sigma_B^{\text{reg}})^{-1/2},
\label{eq:bw_map}
\end{equation}
mapping $z_b^{\text{aligned}} = \mu_A + T_{B \to A}(z_b - \mu_B)$ (or first-order mean alignment $z_b^{\text{aligned}} = \mu_A + (z_b - \mu_B)$ for capitalization-only shifts) before centering by $\mu_A$ and unit-normalizing onto $\mathbb{S}^{d_z - 1}$.

\paragraph{2. MMSE Wiener Spherical Attenuation Compensation.}
Paraphrased clinical definitions across taxonomies sit at centered top-1 cosine affinity $s_b = \max_{a \in A} \max(\hat{z}_b^\top \hat{z}_a, 0) < 1$, attenuating cosine logits proportionally to $s_b$. Under the spherical signal-plus-noise model $y = s_b u + \sqrt{1 - s_b^2}\,\xi$ with noise-to-signal ratio $\lambda = 1/3$, the closed-form linear MMSE Wiener estimator applies a per-code scalar gain:
\begin{equation}
g_\lambda(s_b) = \frac{s_b}{s_b^2 + \lambda (1 - s_b^2)},
\label{eq:wiener}
\end{equation}
scaling only the cosine interaction term: $\phi_\theta^{\text{W}}(x, \hat{z}_b) = \beta_0 + g_\lambda(s_b)\,\big(\phi_\theta(x, \hat{z}_b) - \beta_0\big)$. This enforces exact fixed-point invariance $g_\lambda(1) = 1$, smooth paraphrase amplification above crossover $s^* = \lambda/(1-\lambda) = 0.50$, and automatic shrinkage below $s^*$.

\paragraph{3. Whitened CSLS Hubness Centering, Cardinality Shift, and Centered Prior Projection.}
Using source ZCA whitening $W_A = (\Sigma_A + \epsilon_w I)^{-1/2}$ ($\epsilon_w = 10^{-4}$), we compute each target code's mean cosine similarity $r_B(b)$ to its $k_{\text{csls}} = 10$ nearest neighbors in $B$~\cite{conneau2018word} and subtract $\alpha_{\text{csls}} (r_B(b) - \bar{r}_A)$ ($\alpha_{\text{csls}} = 0.5$). When multi-label catalog cardinality contracts sharply ($|B| < |A|/10$), maximum-entropy cardinality shift adds $\log(|A|/|B|)$; when $|B| \ge |A|/10$, zero-mean centered semantic prior projection transfers source prevalence:
\begin{equation}
\hat{\delta}_b^{(B)} = \tau_{\text{proj}} \Big( s_b^2 \, \delta_{a^*(b)}^{(A)} - \frac{1}{|B|}\sum_{b' \in B} s_{b'}^2 \, \delta_{a^*(b')}^{(A)} \Big),
\label{eq:prior_proj}
\end{equation}
with $\tau_{\text{proj}} \in [0, \tau]$ and linearly interpolated source-validation decision threshold $\theta(\tau_{\text{proj}}) = (1 - \tau_{\text{proj}}/\tau)\,\theta_{\text{pure}} + (\tau_{\text{proj}}/\tau)\,\theta_{\text{cal}}$.

\paragraph{4. Bounded Compositional Set Algebra and Span-Grounded Semantic NMS.}
For target schemas containing public disjunctive group codes ($G = \bigvee_{m \in M_G} m$), group probabilities are clamped analytically onto $[\max_{m \in M_G} p_m,\; 1 - \prod_{m \in M_G} (1 - p_m)]$ in double-precision log space. During multi-label inference, GPU-vectorized greedy non-maximum suppression iterates over above-threshold codes in descending probability order and suppresses any lower-scoring sibling $b'$ claiming the same or adjacent 4-token span ($|s^*(b') - s^*(b)| \le 1$) whenever $\hat{z}_b^\top \hat{z}_{b'} \ge \cos_{\text{nms}}$ ($\cos_{\text{nms}} = 0.80$).

\paragraph{Metadata-Driven Compilation Policy.}
To eliminate per-target hyperparameter selection, all five target schemas are compiled by a single deterministic rule reading only observable public codebook metadata (task mode, uppercase character ratio, non-Latin script ratio, catalog size $|B|/ |A|$, and presence of public group definitions) with universal fixed constants ($\alpha_{\text{csls}} = 0.5$, $\lambda = 1/3$, $\cos_{\text{nms}} = 0.80$): single-label schemas enable mean alignment when uppercase ratio $\ge 0.70$ alongside CSLS and Wiener gain; multi-label schemas enable CSLS, Wiener gain, Span-NMS, bounded set algebra when group definitions exist, Bures--Wasserstein alignment when non-Latin ratio $\ge 0.30$ or $|B| < |A|/10$, cardinality shift when $|B| < |A|/10$, and centered prior projection ($\tau_{\text{proj}} = 0.5$) when $|B| \ge |A|/10$.

\begin{table*}[t]
\centering
\refstepcounter{algocf}\label{alg:isotoken}
\caption*{{\bfseries Algorithm~\thealgocf.} Day-One Clinical Codebook Compilation and Dual-Process Coding in Laya-IsoToken.}
\footnotesize
\setlength{\tabcolsep}{6pt}
\begin{tabular}{@{}p{0.97\textwidth}@{}}
\toprule
\textbf{Inputs:} Frozen source span reader $\phi_\theta$ with global scalars $(\alpha_{\text{sc}}, \beta_0)$ and validation thresholds $(\theta_{\text{pure}}, \theta_{\text{cal}})$; aggregate source codebook summary $\mathcal{S}_A = (\mu_A, \Sigma_A, \hat{Z}_A, \delta^{(A)}, \bar{r}_A, G_A)$; public target codebook $B = \{(b, t_b)\}_{b=1}^{|B|}$ with optional group rules $\mathcal{G}_B$; offline frozen vectorizer $f_{\text{sem}}$; optional offline LLM verifier $\mathcal{M}_{\text{LLM}}$ and escalation budget $q \in [0, 1]$. \\
\midrule
\textbf{Phase I: One-Time Offline Codebook Compilation ($0.2$--$2.6\,\text{s}$, zero target clinical notes)} \\
\quad 1. \textit{Vectorize public descriptions:} Encode $z_b \leftarrow f_{\text{sem}}(t_b)$ for all $b \in B$ (and pool group centroids if $\mathcal{G}_B \neq \emptyset$). \\
\quad 2. \textit{Pre-deployment extrapolation risk:} Center $\hat{z}_b^{\text{raw}} \leftarrow \operatorname{norm}(z_b - \mu_A)$ and compute $\rho_b \leftarrow \sqrt{\lambda_r (\hat{z}_b^{\text{raw}})^\top (G_A + \lambda_r I)^{-1} \hat{z}_b^{\text{raw}}}$ (Eq.~\ref{eq:rho}). \\
\quad 3. \textit{Distribution alignment:} Inspect public metadata of $B$; set $z_b^{\text{al}} \leftarrow \mu_A + T_{B \to A}(z_b - \mu_B)$ via Eq.~\ref{eq:bw_map} (non-Latin $\ge 0.30$ or $|B| < |A|/10$), $\mu_A + (z_b - \mu_B)$ (single-label uppercase $\ge 0.70$), or $z_b$; normalize $\hat{z}_b \leftarrow \operatorname{norm}(z_b^{\text{al}} - \mu_A)$. \\
\quad 4. \textit{Paraphrase gain \& bias compilation:} Compute nearest source affinity $s_b \leftarrow \max_{a \in A}\max(\hat{z}_b^\top \hat{z}_a, 0)$ and $a^*(b) \leftarrow \arg\max_{a \in A} \hat{z}_b^\top \hat{z}_a$; compile Wiener gain $g_\lambda(s_b)$ (Eq.~\ref{eq:wiener}), whitened CSLS offset $-\alpha_{\text{csls}}(r_B(b) - \bar{r}_A)$, and cardinality/prior bias $b_{\text{cal}}(b)$ (Eq.~\ref{eq:prior_proj}) with threshold $\theta_B$. \\
\quad \textbf{Return compiled artifact} $\mathcal{C}_B = (\hat{Z}_B, g_\lambda, b_{\text{cal}}, \rho, \theta_B, \mathcal{G}_B)$. \\
\midrule
\textbf{Phase II: Millisecond Span Inference \& Risk-Guided Dual-Process Verification (per admission $x$)} \\
\quad 5. \textit{System-1 span encoding:} Encode chunks of $x$, mask special tokens, and pool 4-token spans into unit keys/values $\{(\hat{k}_t, \hat{v}_t)\}_{t=1}^{S}$. \\
\quad 6. \textit{Compiled span-to-concept scoring:} For all $b \in B$, compute $\alpha_{t,b} \leftarrow \operatorname{softmax}_t(\gamma_{\text{attn}} \hat{k}_t^\top \hat{z}_b)$, argmax evidence span $s^*(b) \leftarrow \arg\max_t \alpha_{t,b}$, and compiled logit $\ell_b(x) \leftarrow \beta_0 + \alpha_{\text{sc}}\, g_\lambda(s_b)\sum_{t=1}^{S} \alpha_{t,b}(\hat{v}_t^\top \hat{z}_b) + b_{\text{cal}}(b)$; apply bounded set algebra over $\mathcal{G}_B$ if present. \\
\quad 7. \textit{Decision \& span deduplication:} If single-label, set $\hat{Y}_{\text{S1}} \leftarrow \{\arg\max_b \ell_b(x)\}$; else apply GPU Span-NMS ($\cos_{\text{nms}}=0.80$, window $\pm 1$ span) over $\{b : \sigma(\ell_b(x)) \ge \theta_B\}$ to obtain $\hat{Y}_{\text{S1}}$ and Top-$K$ shortlist $\mathcal{L}_K(x)$. \\
\quad 8. \textit{Selective System-2 verification:} Compute note risk $\bar{\rho}(x) \leftarrow \operatorname{mean}_{b \in \mathcal{L}_K(x)} \rho_b$. If $\bar{\rho}(x)$ ranks within budget $q$, pass full note $x$, $\hat{Y}_{\text{S1}}$, $\mathcal{L}_K(x)$, $\rho_b$, and decoded 4-token spans $x[s^*(b)]$ to offline $\mathcal{M}_{\text{LLM}}$ and restrict output strictly to catalog $B$; else return $\hat{Y}_{\text{S1}}$. \\
\bottomrule
\end{tabular}
\end{table*}

\subsection{Subspace Extrapolation Residual and Dual-Process Verification Cascade}

Adapting closed-form subspace geometry~\cite{heyden2026cast}, we form the scale-invariant centered source Gram matrix $G_A = \frac{d_z}{|A|} \hat{Z}_A^\top \hat{Z}_A$ on unwhitened unit-normalized source embeddings ($\operatorname{tr}(G_A) = d_z$) and ridge projection $P_A = G_A (G_A + \lambda_r I)^{-1}$ ($\lambda_r = 1.0$). The pre-deployment extrapolation residual of target code $b$ is:
\begin{equation}
\rho_b = \sqrt{\lambda_r \, (\hat{z}_b^{\text{raw}})^\top (G_A + \lambda_r I)^{-1} \hat{z}_b^{\text{raw}}} = \big\| (I - P_A)^{1/2} \hat{z}_b^{\text{raw}} \big\|_2 \in [0, 1].
\label{eq:rho}
\end{equation}

For dual-process verification~\cite{kahneman2011thinking,soroush2024llm,edin2026symphony} (Fig.~\ref{fig:framework}c), we pair System-1 Laya-IsoToken with two offline open-weight instruction models spanning two parameter scales—\texttt{Qwen/Qwen3.5-9B}~\cite{qwen2026qwen35} ($n=300$, $295$ unseen patients) and \texttt{google/gemma-4-31B-it}~\cite{google2026gemma4} ($n=1{,}000$, $935$ unseen patients; greedy decoding, \texttt{max\_new\_tokens} $= 512$)—on consecutive patient-disjoint locked MIMIC-IV ICD-10 test admissions. Standalone and cascade arms receive identical, untruncated discharge summaries (up to $30{,}000$ characters, covering $100\%$ of 4,000-word notes). We evaluate four pre-specified modes: (i) Standalone zero-shot LLM coding; (ii) System-1 Laya-IsoToken alone; (iii) Full Audit Cascade, where System-2 reviews System-1's above-threshold proposals alongside remaining Top-50 shortlist candidates with official descriptions, probabilities, $\rho_b$, and 4-token spans, plus a zero-tuning Confidence-Anchored Cascade retaining high-confidence System-1 anchors ($p \ge 0.50$); and (iv) Selective Escalation across budgets $q \in \{10\%, 20\%, 30\%, 50\%, 75\%\}$ prioritized by note-level mean candidate $\rho_b$, benchmarked against 50-replicate random routing.

\subsection{Baselines and Statistical Analysis}

We benchmark Laya-IsoToken against published same-split supervised coders~\cite{mullenbach2018explainable,li2020multirescnn,vu2020label,huang2022plmicd,edin2023automated,liu2026duallaat,cossin2020iam} and seven zero-target-training baselines: (1) BM25 lexical retrieval~\cite{robertson2009bm25} calibrated on source ICD-9 validation; (2--4) dense bi-encoder retrieval using general multilingual \texttt{Qwen3-Embedding-0.6B}~\cite{zhang2025qwen3embedding}, dual-encoder biomedical \texttt{MedCPT}~\cite{jin2023medcpt}, and ontology \texttt{BioLORD-2023-M}~\cite{remy2024biolord} with identical alignment operators; and architecture-matched 22-layer \texttt{Laya-PLM-ICD} ($17.4$M trainable label parameters) transferred via (5) norm-matched CAST~\cite{heyden2026cast}, (6) official 2018 US CMS General Equivalence Mappings (forward, bidirectional, and bidirectional + Span-NMS)~\cite{boyd2013discriminatory}, and (7) top-$5$ Semantic Crosswalk with metadata-driven calibration and Span-NMS.

Models are trained on $4 \times$ NVIDIA GH200 GPUs in \texttt{bfloat16} distributed data-parallel mode using AdamW (learning rate $5 \times 10^{-5}$ for transformer layers and $10^{-3}$ for projection parameters, $5\%$ warmup, cosine decay, effective batch size $64$, patience $3$ on source validation Micro-F1) across three random seeds ($42, 43, 44$). Hypothesis testing uses $B = 2{,}000$ patient-clustered (or document-clustered on CodiEsp) paired bootstrap replicates with two-sided 95\% confidence intervals and finite-sample Phipson--Smyth two-sided $p$-values. Alongside corpus-level score differences, we report three standardized per-patient paired effect sizes~\cite{cohen2013statistical}: paired Cohen's $d_z = \bar{\Delta}_u / s_{\Delta_u}$ over patient-level F1 (or accuracy) differences $\Delta_u$, paired rank-biserial correlation $r_{\text{rb}} = (W - L)/(W + L)$, and non-tie patient win rate $W/(W + L)$, where $W$ and $L$ count patients on which method A strictly outperforms or trails method B.

\section*{Data Availability}
All clinical datasets analyzed in this study are accessible publicly or under credentialed access from their original repositories. MIMIC-IV v2.2 clinical discharge summaries, ICD-9-CM, ICD-10-CM/PCS, and MS-DRG records are available via PhysioNet (\url{https://physionet.org/content/mimiciv/2.2/} and \url{https://physionet.org/content/mimic-iv-note/2.2/}) upon completion of credentialed access and CITI training~\cite{johnson2023mimic,johnson2023mimicnote}. The Spanish CLEF eHealth 2020 CodiEsp corpus and gold test annotations are publicly distributed on Zenodo (records 3693570 and 3837305)~\cite{miranda2020codiesp}. The Greek ElCardioCC BioASQ cardiology coding corpus~\cite{dimitriadis2025elcardiocc} and Chinese CBLUE CHIP-CDN clinical diagnosis normalization benchmark~\cite{zhang2022cblue} are available from their respective shared-task organizers. Public 2018 US CMS General Equivalence Mappings~\cite{boyd2013discriminatory}, CMS MS-DRG definitions (v25--v37), and WHO ICD-10 terminology releases are openly downloadable from CMS and WHO. Deterministic cohort preparation scripts regenerating the exact patient-disjoint splits from credentialed PhysioNet files without redistributing protected health information are included in the public repository.

\section*{Code Availability}
All core source code for Laya-IsoToken training, offline closed-form codebook compilation, pre-deployment subspace extrapolation auditing, 4-token span deletion faithfulness evaluation, full-note dual-process LLM verification cascades, and patient-clustered paired bootstrap and standardized effect-size analysis is openly available at \url{https://github.com/ZeqiangWangAI/Laya-IsoToken} under the GNU Affero General Public License v3.0 (\texttt{AGPL-3.0-only}).

\section*{Ethics Statement}
This study analyzed exclusively de-identified secondary clinical records governed by established institutional review board exemptions and PhysioNet data use agreements (MIMIC-IV v2.2) alongside publicly released de-identified multilingual shared-task corpora (CodiEsp, ElCardioCC, CHIP-CDN). No re-identification was attempted and no protected health information was transmitted to external commercial APIs; all neural training, embedding extraction, and large language model verification ran strictly offline inside a private high-performance computing environment.

\section*{Competing Interests}
The author declares no competing financial or non-financial interests.

\bibliographystyle{unsrtnat}
\bibliography{references}

\clearpage
\section*{Supplementary Information}
\renewcommand{\thetable}{S\arabic{table}}
\setcounter{table}{0}
\renewcommand{\thefigure}{S\arabic{figure}}
\setcounter{figure}{0}
Supplementary Tables~\ref{tab:subgroups}--\ref{tab:effect_sizes} and Supplementary Figs.~\ref{fig:stage8_diag}--\ref{fig:encoder_scaling} report TRIPOD+AI demographic and clinical-service subgroup fairness, 21-chapter ICD-10-CM/PCS error decomposition, six-benchmark cohort reconciliation, locked ICD-10 probability calibration progression, Wiener spherical gain sensitivity, public fan-out-gated CMS GEMs procedure hybridization, MS-DRG structured severity headroom, lexical-neural hybrid dispatch, System-2 decision-transition audits, five-encoder offline codebook scaling across $384$d--$2560$d, and complete patient-clustered paired hypothesis tests with standardized effect sizes.

\begin{table*}[ht!]
\centering
\caption{TRIPOD+AI subgroup audit on the locked patient-disjoint MIMIC-IV ICD-10 test cohort ($n=13{,}783$ admissions, $8{,}255$ unseen patients; median note length $1{,}584$ words, IQR $1{,}206$--$2{,}091$; median $14$ codes per admission, IQR $9$--$20$). Laya-IsoToken (Qwen3 codebook, seed 42, full operators) shows near-zero sex disparity ($0.22\%$ gap) and outperforms Laya-PLM-ICD + Semantic Crosswalk + NMS across every clinical service, note length quartile, and comorbidity burden quartile.}
\label{tab:subgroups}
\footnotesize
\setlength{\tabcolsep}{4.5pt}
\begin{tabular}{@{}lrrcccc@{}}
\toprule
Subgroup Stratum & Admissions & Patients & Laya-IsoToken & Crosswalk & $\Delta$ F1 & Laya-IsoToken P / R \\
\midrule
\multicolumn{7}{@{}l}{\textit{Sex}} \\
F & 6,883 & 4,167 & 32.16 & 27.23 & +4.93 & 37.1 / 28.4 \\
M & 6,900 & 4,088 & 32.38 & 28.10 & +4.28 & 36.2 / 29.3 \\
\midrule
\multicolumn{7}{@{}l}{\textit{Clinical Service}} \\
Medicine & 7,529 & 4,427 & 35.66 & 30.96 & +4.70 & 39.3 / 32.6 \\
OB/GYN & 482 & 394 & 21.07 & 14.15 & +6.92 & 32.9 / 15.5 \\
Orthopaedics & 868 & 701 & 20.03 & 16.22 & +3.81 & 23.6 / 17.4 \\
Other & 115 & 100 & 26.50 & 23.67 & +2.84 & 29.9 / 23.8 \\
Psychiatry & 348 & 300 & 34.73 & 26.49 & +8.25 & 40.9 / 30.2 \\
Surgery & 4,441 & 3,449 & 27.19 & 22.80 & +4.39 & 32.4 / 23.4 \\
\midrule
\multicolumn{7}{@{}l}{\textit{Discharge Note Length Quartile}} \\
Q1 (shortest) & 3,447 & 2,850 & 28.47 & 23.53 & +4.94 & 34.0 / 24.5 \\
Q2 & 3,457 & 2,853 & 31.78 & 26.83 & +4.95 & 36.5 / 28.2 \\
Q3 & 3,435 & 2,688 & 33.69 & 28.78 & +4.91 & 37.9 / 30.3 \\
Q4 (longest) & 3,444 & 2,461 & 33.19 & 29.19 & +3.99 & 36.9 / 30.1 \\
\midrule
\multicolumn{7}{@{}l}{\textit{Comorbidity Burden Quartile}} \\
Q1 ($\le$ 9 codes) & 3,960 & 3,273 & 29.17 & 24.62 & +4.55 & 30.7 / 27.7 \\
Q2 (10-14) & 3,365 & 2,696 & 32.58 & 27.01 & +5.57 & 35.9 / 29.8 \\
Q3 (15-20) & 3,169 & 2,369 & 34.09 & 29.04 & +5.05 & 38.4 / 30.6 \\
Q4 (> 20 codes) & 3,289 & 2,225 & 31.93 & 28.02 & +3.91 & 37.8 / 27.6 \\
\bottomrule
\end{tabular}
\end{table*}

\begin{table*}[ht!]
\centering
\caption{Per-chapter ICD-10-CM/PCS breakdown on the locked patient-disjoint MIMIC-IV ICD-10 test cohort ($n=13{,}783$, seed 42). Zero-training Laya-IsoToken outperforms Laya-PLM-ICD + Semantic Crosswalk + NMS on 20 of 21 clinical chapters, while pinpointing structural failure modes in Injury/Trauma ($7.76\%$; External Causes $11.00\%$), Obstetrics ($8.93\%$), and 7-character ICD-10-PCS procedures ($2.93\%$). On MS-DRG ($n=11{,}869$), collapsing CC/MCC billing severity tiers into base clinical families lifts seed-42 Top-1 accuracy from $6.21\%$ to $14.49\%$ for Laya-IsoToken (pplx codebook; $10.71\%$ 3-seed mean for Qwen3 codebook) versus $8.45\%$ for Crosswalk.}
\label{tab:chapters}
\scriptsize
\setlength{\tabcolsep}{4.5pt}
\begin{tabular}{@{}lrrccc@{}}
\toprule
ICD-10 Clinical Chapter & Gold Instances & Unique Codes & Laya-IsoToken & Crosswalk & $\Delta$ F1 \\
\midrule
Health Status/Factors (Z) & 31,178 & 544 & 37.73 & 36.83 & +0.90 \\
Circulatory (I) & 27,576 & 664 & 27.02 & 17.54 & +9.48 \\
ICD-10-PCS Procedures & 23,576 & 3,987 & 2.93 & 1.38 & +1.55 \\
Endocrine/Metabolic (E) & 22,506 & 259 & 46.24 & 43.34 & +2.91 \\
Symptoms/Signs (R) & 13,813 & 412 & 35.54 & 32.16 & +3.39 \\
Digestive (K) & 12,827 & 479 & 34.03 & 25.59 & +8.44 \\
Mental/Behavioral (F) & 11,168 & 304 & 31.70 & 14.66 & +17.04 \\
Genitourinary (N) & 8,468 & 287 & 57.33 & 58.53 & -1.20 \\
External Causes (V-Y) & 8,321 & 410 & 11.00 & 6.31 & +4.69 \\
Blood/Immune (D50-D89) & 8,063 & 139 & 47.04 & 41.81 & +5.23 \\
Nervous System (G) & 7,850 & 318 & 47.46 & 46.97 & +0.50 \\
Respiratory (J) & 7,646 & 213 & 30.12 & 24.42 & +5.70 \\
Injury/Trauma (S-T) & 7,630 & 1,812 & 7.76 & 3.81 & +3.94 \\
Musculoskeletal (M) & 7,038 & 738 & 21.36 & 17.69 & +3.67 \\
Neoplasms (C-D49) & 5,431 & 504 & 33.59 & 24.66 & +8.93 \\
Infectious (A-B) & 4,211 & 214 & 38.56 & 18.40 & +20.16 \\
Skin/Subcutaneous (L) & 2,484 & 316 & 16.66 & 10.00 & +6.66 \\
Eye/Ear (H) & 1,650 & 284 & 35.37 & 30.03 & +5.34 \\
Pregnancy/Obstetrics (O) & 1,120 & 337 & 8.93 & 3.89 & +5.04 \\
Congenital (Q) & 509 & 132 & 27.22 & 26.39 & +0.83 \\
Perinatal (P) & 4 & 2 & 33.33 & 0.00 & +33.33 \\
\bottomrule
\end{tabular}
\end{table*}

\clearpage
\begin{table*}[ht!]
\centering
\caption{Cohort, split, and patient-disjointness reconciliation across all six clinical benchmarks. Deterministic patient-disjoint cohort scripts (\texttt{scripts/prepare\_locked\_test\_splits.py}) regenerate exact splits directly from credentialed PhysioNet MIMIC-IV v2.2 and public Edin/DualLAAT benchmark IDs with zero clinical text redistribution.}
\label{tab:cohorts}
\scriptsize
\setlength{\tabcolsep}{3.6pt}
\resizebox{\textwidth}{!}{%
\begin{tabular}{@{}lclrrll@{}}
\toprule
Benchmark Split & Lang & Catalog Size & Admissions & Unique Patients & Overlap Control & Evaluation Role \\
\midrule
MIMIC-IV ICD-9 Source Train & EN & 11,324 & 154,542 & 72,216 & 0 & Source training \\
MIMIC-IV ICD-9 Source Val & EN & 11,324 & 21,980 & 10,068 & 0 & Model/threshold selection \\
MIMIC-IV ICD-9 Locked Test & EN & 11,324 (6,150 freq.) & 32,804 & 15,425 & 0 & Locked in-domain test \\
\midrule
MIMIC-IV ICD-10 Dev (disjoint) & EN & 26,085 & 9,077 & 5,394 & 4,301 excl. & Dev ablation only \\
MIMIC-IV ICD-10 Locked Test (disjoint) & EN & 26,085 & 13,783 & 8,255 & 6,019 excl. & Locked zero-train test \\
MIMIC-IV ICD-10 Cascade-300 / 1,000 & EN & 26,085 & 300 / 1,000 & 295 / 935 & Disjoint subset & Locked System-2 audit \\
MIMIC-IV MS-DRG Dev (disjoint) & EN & 1,581 & 7,875 & 4,779 & Source-disjoint & Dev diagnostics only \\
MIMIC-IV MS-DRG Locked Test (disjoint) & EN & 1,581 & 11,869 & 7,340 & Source-disjoint & Locked zero-train test \\
CLEF 2020 CodiEsp Dev / Locked Test & ES & 185,458 (+11,381 grp) & 250 / 250 & 250 / 250 & Independent & Dev / Locked gold test \\
BioASQ ElCardioCC Train & EL/EN & 115 & 2,500 & 2,500 & Independent & Non-test boundary (test hidden) \\
CBLUE CHIP-CDN Dev & ZH & 37,645 & 2,000 & 2,000 & Independent & Non-test boundary (test hidden) \\
\bottomrule
\end{tabular}%
}
\end{table*}

\begin{table*}[ht!]
\centering
\caption{Supplementary calibration, zero-target public fan-out-gated CMS GEMs procedure hybrid, MS-DRG structured CC/MCC headroom, and zero-target lexical-neural hybrid audits. Panel~A details the accuracy-calibration tradeoff across closed-form operators on locked ICD-10; Panel~B confirms that restricting CMS GEMs to deterministic low-branching surgical mappings ($\text{fan-out}\le 2$) raises zero-training ICD-10 PCS F1 $4.8\times$ ($3.05\% \to 14.66\%$) and overall Micro~F1 to $32.84 \pm 0.28\%$ ($p<0.0005$ on 3/3 seeds); Panel~C quantifies how MS-DRG text accuracy decays $2.6\times$ across comorbidity tiers and $1.9\times$ for post-acute dispositions; Panel~D validates zero-target-tuning lexical-neural dispatch across all five benchmarks (lifting Spanish CodiEsp to $12.94\%$ and doubling shortlist R@50 from $12.6\%$ to $24.8\%$, while preserving English ICD-10 at $32.28\%$ and cross-script ElCardioCC).}
\label{tab:cal_hybrids}
\scriptsize
\setlength{\tabcolsep}{3.6pt}
\resizebox{\textwidth}{!}{%
\begin{tabular}{@{}lcccccc@{}}
\toprule
Operating Point / Audit Stratum & MiF1 / Top-1 & MiAUC / Base & MiAUPRC / PCS / Gap & Active ECE / $n$ & Active Brier & Latency / Note \\
\midrule
\multicolumn{7}{@{}l}{\textit{Panel A: Locked MIMIC-IV ICD-10 calibration progression ($n=13{,}783$, Qwen3 codebook, seed 42, GPU Span-NMS)}} \\
Raw cosine (0 operators) & 25.17 & 97.17 & 15.57 & 0.0537 & 0.1219 & 9.07\,ms \\
+ CSLS hubness centering ($\alpha_{\text{csls}}=0.5$) & 25.72 & 97.16 & 16.22 & 0.0473 & 0.1247 & 9.02\,ms \\
+ Centered Bayesian Prior Proj. ($\tau_{\text{proj}}=0.5$) & 30.35 & 97.43 & 20.88 & 0.1192 & 0.1210 & 9.06\,ms \\
+ Full Compiled Policy (Wiener $\lambda=1/3$ + Span-NMS) & 32.27 & 97.42 & 19.91 & 0.1341 & 0.1227 & 9.41\,ms \\
\midrule
\multicolumn{7}{@{}l}{\textit{Panel B: Zero-target public fan-out-gated CMS GEMs PCS hybrid on locked ICD-10 test (3 seeds, $n=13{,}783$)}} \\
Laya-IsoToken Alone (zero crosswalk) & 32.20 $\pm$ 0.26 & --- & 3.05 $\pm$ 0.14 (PCS) & --- & --- & Baseline \\
Naive unconstrained CMS GEMs PCS swap (fan-out $\le 453$) & 28.54 $\pm$ 0.49 & --- & 8.25 $\pm$ 0.16 (PCS) & --- & --- & +88k PCS FPs \\
Public 1-to-1 CMS GEMs PCS hybrid (fan-out $= 1$) & 32.64 $\pm$ 0.28 & --- & 10.41 $\pm$ 0.97 (PCS) & --- & --- & 3/3 seeds $p<0.0005$ \\
Public low-branch CMS GEMs PCS hybrid (fan-out $\le 2$) & 32.84 $\pm$ 0.28 & --- & 14.66 $\pm$ 1.04 (PCS) & --- & --- & 3/3 seeds $p<0.0005$ \\
\midrule
\multicolumn{7}{@{}l}{\textit{Panel C: Locked MS-DRG ($n=11{,}869$, pplx s42) structured CC/MCC headroom by comorbidity \& disposition}} \\
Q1 ($\le$8 dx) & 9.32 (Exact) & 18.67 (Base) & +9.35 (Gap) & 3,079 adm. & --- & --- \\
Q2 (9-13 dx) & 6.38 (Exact) & 14.58 (Base) & +8.20 (Gap) & 3,072 adm. & --- & --- \\
Q3 (14-19 dx) & 5.29 (Exact) & 12.05 (Base) & +6.76 (Gap) & 2,929 adm. & --- & --- \\
Q4 ($\ge$20 dx) & 3.55 (Exact) & 8.82 (Base) & +5.27 (Gap) & 2,789 adm. & --- & --- \\
Discharge: Home & 7.92 (Exact) & 16.64 (Base) & +8.71 (Gap) & 5,326 adm. & --- & --- \\
Discharge: Home with Service & 5.28 (Exact) & 11.18 (Base) & +5.90 (Gap) & 3,201 adm. & --- & --- \\
Discharge: Post-Acute Facility & 4.23 (Exact) & 11.30 (Base) & +7.07 (Gap) & 2,601 adm. & --- & --- \\
Discharge: Expired/Died & 5.15 (Exact) & 9.54 (Base) & +4.38 (Gap) & 388 adm. & --- & --- \\
\midrule
\multicolumn{7}{@{}l}{\textit{Panel D: Zero-target-tuning BM25 + Laya-IsoToken lexical-neural hybrid (seed 42; source-val BM25 $\theta=0.85$)}} \\
MIMIC ICD-10 Locked Test (EN narrative) & 32.28 (Hybrid) & 32.28 (Neural) & 5.71 (BM25) & 52.1\% $\to$ 45.0\% (R@50) & --- & EN neural \\
MIMIC MS-DRG Locked Test (EN narrative) & 6.21 (Hybrid) & 6.21 (Neural) & 0.91 (BM25) & 55.6\% $\to$ 53.2\% (R@50) & --- & EN neural \\
CLEF 2020 Spanish CodiEsp Locked Test (ES) & 12.94 (Hybrid) & 5.23 (Neural) & 12.76 (BM25) & 12.6\% $\to$ 24.8\% (R@50) & --- & BM25 $\cup$ $p\ge 0.50$ \\
BioASQ Greek ElCardioCC (EL $\to$ EN) & 13.33 (Hybrid) & 13.33 (Neural) & 0.00 (BM25) & 62.7\% $\to$ 62.7\% (R@50) & --- & EN neural \\
CBLUE Chinese CHIP-CDN (ZH short phrase) & 35.62 (Hybrid) & 0.86 (Neural) & 35.62 (BM25) & 10.3\% $\to$ 69.3\% (R@50) & --- & Short-phrase BM25 \\
\bottomrule
\end{tabular}%
}
\end{table*}

\begin{figure*}[ht!]
    \centering
    \includegraphics[width=\textwidth]{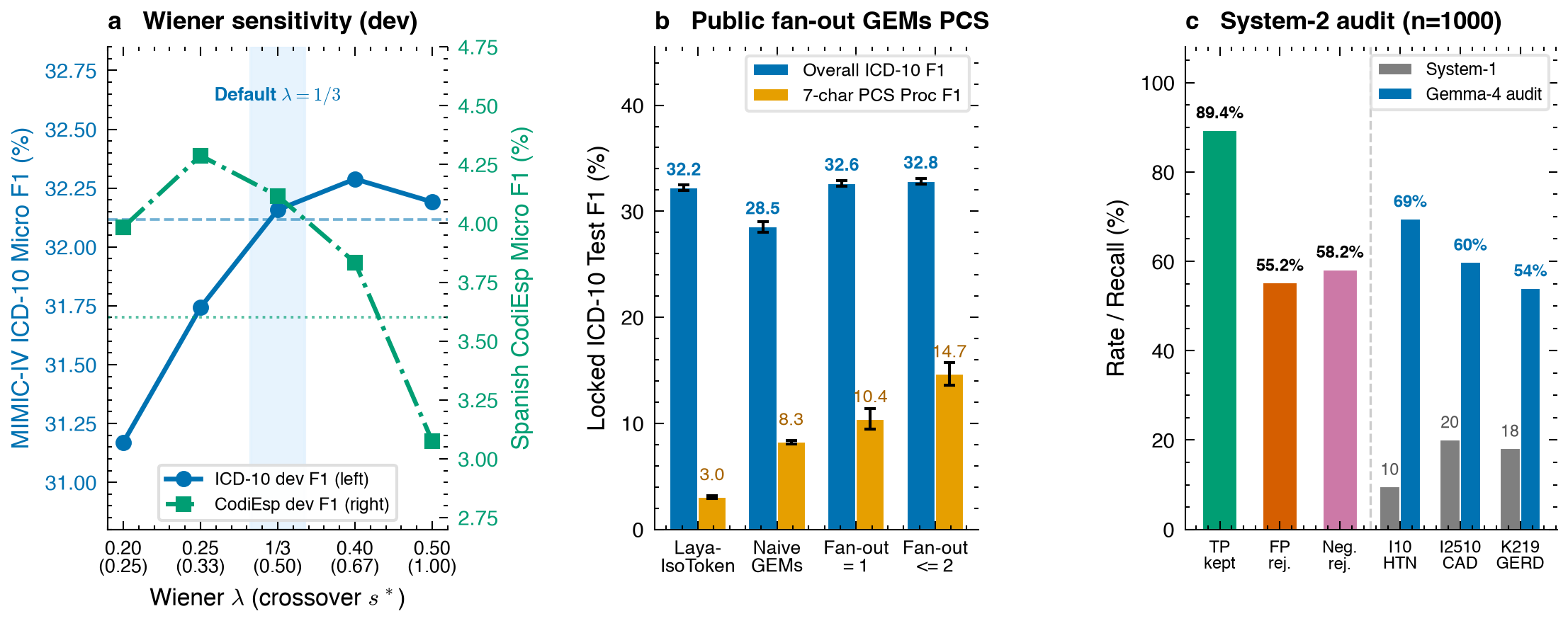}
    \caption{Supplementary mechanistic and deployment-hybrid diagnostics. \textbf{a}, Six-point MMSE Wiener noise-to-signal ($\lambda$) sensitivity across MIMIC-IV ICD-10 dev ($n=9{,}077$) and Spanish CodiEsp dev ($n=250$) with GPU Span-NMS, verifying $\lambda=1/3$ ($s^*=0.50$) as the joint cross-lingual operating point. \textbf{b}, Three-seed public fan-out-gated CMS GEMs procedure hybrid on locked MIMIC-IV ICD-10 test ($n=13{,}783$): unconstrained GEMs substitution degrades overall F1 ($32.20\% \to 28.54\%$), whereas public low-branching surgical links ($\text{fan-out}\le 2$) raise PCS F1 $4.8\times$ ($3.05\% \to 14.66\%$) and overall F1 to $32.84 \pm 0.28\%$ ($p<0.001$). \textbf{c}, System-2 \texttt{Gemma-4-31B-it} decision-transition audit across all $n=1{,}000$ locked ICD-10 admissions ($935$ unseen patients), showing retention of $89.4\%$ of System-1 true positives, rejection of $55.2\%$ of false positives ($58.2\%$ in negated windows), and high-prevalence paraphrase recovery (\texttt{I10} $7.3\times$, \texttt{I2510} $3.0\times$, \texttt{K219} $3.0\times$).}
    \label{fig:stage8_diag}
\end{figure*}

\clearpage
\begin{table*}[ht!]
\centering
\caption{Offline codebook vectorizer scaling across five frozen encoders and four embedding dimensions ($d_z \in \{384, 768, 1024, 2560\}$) under the identical clinical span reader (seed 42, zero trainable label parameters, metadata-driven compilation). Scaling solely the offline codebook vectorizer from \texttt{Qwen3-0.6B} (1024d) to \texttt{Qwen3-4B} (2560d) yields statistically significant gains across every official locked test benchmark: MIMIC-IV ICD-10 rises $+0.73\%$ ($32.28\% \to 33.01\%$, deconfounded LLR $28.10\% \to 30.17\%$, PCS $2.93\% \to 4.49\%$, $p<0.001$), MS-DRG nearly doubles to $8.80\%$ exact Top-1 ($17.95\%$ base-family, $63.45\%$ R@50; $+4.22\%$ vs.\ Qwen3-0.6B, paired $d_z=+0.15$, $p<0.001$), Spanish CodiEsp rises $+1.63\%$ to $6.86\%$ (paired $d_z=+0.19$, $p=0.002$), and source ICD-9 Edin-6,150 Macro~AUC reaches $97.57\%$ at $8.06$\,ms/note.}
\label{tab:encoder_scaling}
\scriptsize
\setlength{\tabcolsep}{2.4pt}
\resizebox{\textwidth}{!}{%
\begin{tabular}{@{}lccccccccccc@{}}
\toprule
Offline Codebook Vectorizer & $d_z$ & Lang & Syn8 & ICD-9 MiF1 & ICD-9 MaAUC & ICD-10 MiF1 & ICD-10 Pure & MS-DRG (Base) & ES Test & EL & ZH \\
\midrule
Self-encoded reader (frozen mmBERT mean-pool) & 768 & Multi & 0.42 & --- & --- & 11.12 (dev) & --- & 0.27 (dev) & 0.03 (dev) & 11.78 & 0.02 \\
\midrule
all-MiniLM-L6-v2 (22M, EN-only)~\cite{wang2020minilm} & 384 & EN & 0.83 & 62.19 & 96.92 & 31.00 & 27.82 & 6.07 (12.9) & 2.27 & 15.42 & 0.00 \\
Qwen3-Embedding-0.6B (causal)~\cite{zhang2025qwen3embedding} & 1024 & Multi & 0.82 & 62.26 & 97.39 & 32.28 & 28.10 & 4.58 (10.0) & 5.23 & 13.33 & 0.86 \\
pplx-embed-v1-0.6B (bidir.)~\cite{perplexity2026pplxembed} & 1024 & Multi & 0.62 & 62.62 & 97.41 & 30.12 & 26.47 & 6.21 (14.5) & 4.23 & 18.85 & 0.78 \\
\rowcolor{tablehl} Qwen3-Embedding-4B (causal)~\cite{zhang2025qwen3embedding} & 2560 & Multi & 0.80 & 62.50 & \textbf{97.57} & \textbf{33.01} & \textbf{30.17} & \textbf{8.80 (18.0)} & \textbf{6.86} & 13.46 & 0.94 \\
\bottomrule
\end{tabular}%
}
\end{table*}

\begin{table*}[ht!]
\centering
\caption{Patient-clustered paired hypothesis tests and standardized effect sizes across locked test benchmarks ($B=2{,}000$ bootstrap replicates; Phipson--Smyth finite-sample two-sided $p$-values). Resampling clusters admissions by unique \texttt{subject\_id} where available (and by clinical document on CodiEsp). Paired Cohen's $d_z = \bar{\Delta}_u / s_{\Delta_u}$~\cite{cohen2013statistical}, paired rank-biserial correlation $r_{\text{rb}} = (W - L)/(W + L)$, and non-tie win rate $W/(W+L)$ quantify per-patient clinical effect magnitude alongside corpus-level Micro~F1 / Top-1 differences. For 3-seed comparisons, $\Delta$, $d_z$, $r_{\text{rb}}$, and win rate report the 3-seed mean, 95\% CI reports seed 42, and $p$ reports the conservative maximum across seeds 42--44.}
\label{tab:effect_sizes}
\scriptsize
\setlength{\tabcolsep}{3.4pt}
\resizebox{\textwidth}{!}{%
\begin{tabular}{@{}llcccccc@{}}
\toprule
Benchmark (Clusters) & Paired Comparison & $\Delta$ Score & 95\% Bootstrap CI & Two-sided $p$ & Cohen's $d_z$ & $r_{\text{rb}}$ & Win Rate \\
\midrule
MIMIC-IV ICD-9 Test ($15{,}425$ pat.) & Laya-IsoToken (Qwen3, 3s) vs.\ Laya-PLM-ICD & +0.50\% & $[+0.47, +0.64]$ & $<0.001$ & +0.04 & +0.07 & 53.4\% \\
MIMIC-IV ICD-9 Test ($15{,}425$ pat.) & Laya-IsoToken (pplx, 3s) vs.\ Laya-PLM-ICD & +0.74\% & $[+0.84, +1.01]$ & $<0.001$ & +0.08 & +0.10 & 55.2\% \\
MIMIC-IV ICD-10 Test ($8{,}255$ pat.) & Laya-IsoToken (Qwen3, 3s) vs.\ ZeroShot BioLORD-2023-M & +29.95\% & $[+29.70, +30.34]$ & $<0.001$ & +1.80 & +0.96 & 98.0\% \\
MIMIC-IV ICD-10 Test ($8{,}255$ pat.) & Laya-IsoToken (Qwen3, 3s) vs.\ Semantic Crosswalk & +5.36\% & $[+4.39, +4.80]$ & $<0.001$ & +0.49 & +0.45 & 72.4\% \\
MIMIC-IV ICD-10 Test ($8{,}255$ pat.) & Laya-IsoToken (Qwen3, 3s) vs.\ Official CMS GEMs+NMS & +3.53\% & $[+3.32, +3.86]$ & $<0.001$ & +0.07 & +0.11 & 55.5\% \\
MIMIC-IV ICD-10 Test ($8{,}255$ pat.) & Laya-IsoToken (Qwen3-4B vs.\ 0.6B, s42) & +0.73\% & $[+0.56, +0.90]$ & $<0.001$ & +0.02 & +0.03 & 51.7\% \\
MIMIC-IV MS-DRG Test ($7{,}340$ pat.) & Laya-IsoToken (pplx, 3s) vs.\ Semantic Crosswalk & +2.00\% & $[+1.33, +2.34]$ & $<0.001$ & +0.06 & +0.25 & 62.4\% \\
MIMIC-IV MS-DRG Test ($7{,}340$ pat.) & Laya-IsoToken (Qwen3-4B vs.\ 0.6B, s42) & +4.22\% & $[+3.65, +4.79]$ & $<0.001$ & +0.15 & +0.44 & 72.1\% \\
Spanish CodiEsp Test ($250$ cases) & Laya-IsoToken (Qwen3, 3s) vs.\ Semantic Crosswalk & +3.92\% & $[+2.54, +4.17]$ & $<0.001$ & +0.54 & +0.72 & 86.2\% \\
Spanish CodiEsp Test ($250$ cases) & Locked Laya-IsoToken vs.\ Unlocked (3s) & +0.86\% & $[-0.47, +1.21]$ & $0.3778$ & -0.02 & -0.04 & 48.1\% \\
Spanish CodiEsp Test ($250$ cases) & Laya-IsoToken (Qwen3-4B vs.\ 0.6B, s42) & +1.63\% & $[+0.63, +2.63]$ & $0.0020$ & +0.19 & +0.23 & 61.4\% \\
ICD-10 Cascade ($935$ pat., $n=1{,}000$) & Gemma-4-31B Cascade vs.\ System-1 Alone & +8.16\% & $[+7.54, +8.77]$ & $<0.001$ & +0.66 & +0.59 & 79.4\% \\
ICD-10 Cascade ($935$ pat., $n=1{,}000$) & Gemma-4-31B Cascade vs.\ Standalone 31B & +9.62\% & $[+8.69, +10.56]$ & $<0.001$ & +0.48 & +0.41 & 70.7\% \\
ICD-10 Cascade ($935$ pat., $n=1{,}000$) & System-1 Alone vs.\ Standalone 31B & +1.47\% & $[+0.49, +2.46]$ & $0.0070$ & -0.02 & +0.02 & 51.2\% \\
\bottomrule
\end{tabular}%
}
\end{table*}

\begin{figure*}[ht!]
    \centering
    \includegraphics[width=\textwidth]{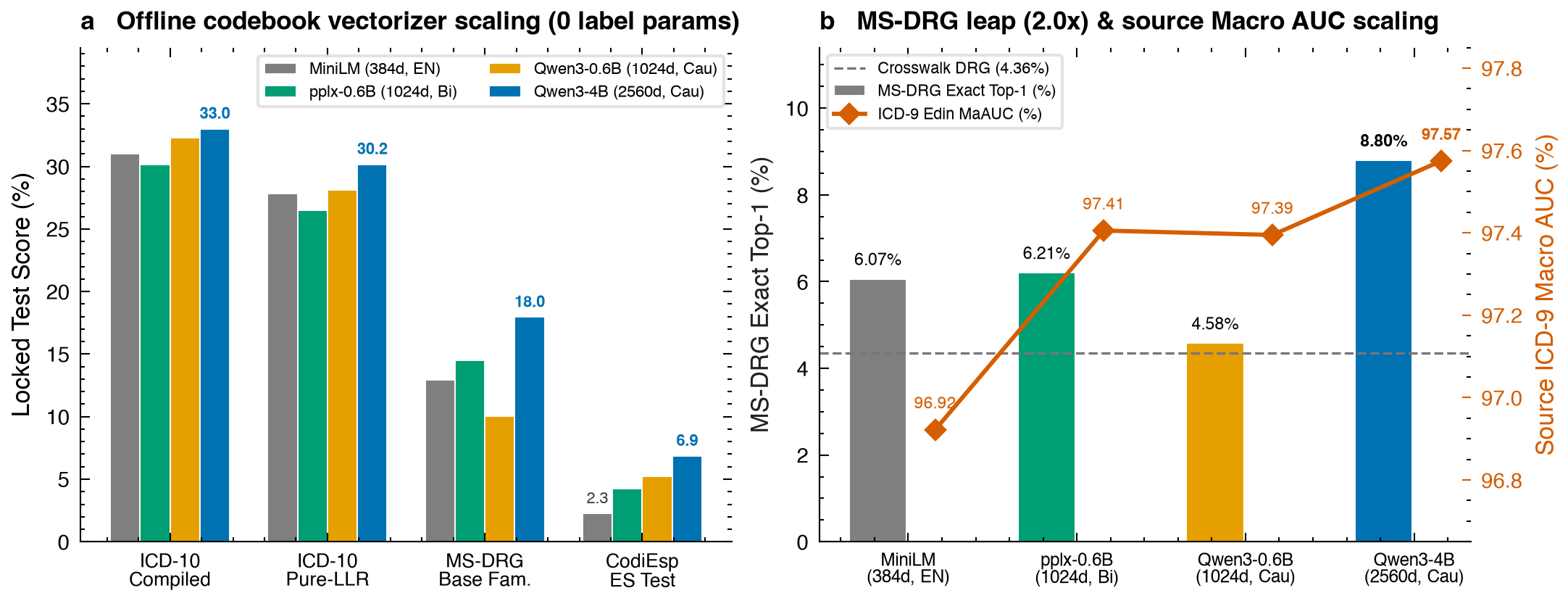}
    \caption{Offline codebook vectorizer scaling across embedding dimensions and model families ($d_z \in \{384, 1024, 2560\}$, zero trainable label parameters, identical clinical span reader, seed 42). \textbf{a}, Locked zero-training transfer across MIMIC-IV ICD-10 (compiled and deconfounded LLR), MS-DRG base clinical families, and Spanish CodiEsp gold test, illustrating monotonic scaling with \texttt{Qwen3-Embedding-4B} and the expected cross-lingual collapse of English-only \texttt{all-MiniLM-L6-v2} on Spanish ($2.27\%$). \textbf{b}, Exact MS-DRG Top-1 accuracy (nearly doubling to $8.80\%$ with \texttt{Qwen3-4B}, $2.0\times$ Semantic Crosswalk at $4.36\%$) paired with locked source MIMIC-IV ICD-9 Edin-6,150 Macro ROC-AUC (climbing monotonically to $97.57\%$).}
    \label{fig:encoder_scaling}
\end{figure*}

\end{document}